\documentclass{IEEEtaes}

\usepackage{color,array,amsthm}
\usepackage{graphicx}
\usepackage{mathrsfs}
\usepackage{enumerate}
\newtheorem{thm}{Theorem}[section]
\usepackage{amsthm,amsmath,amssymb}
\usepackage{multirow,boldline}
\usepackage{tabularx}
\newcolumntype{C}{>{\centering\arraybackslash}X}
\usepackage[linesnumbered,ruled]{algorithm2e}
\usepackage[compatible]{algpseudocode}
\usepackage{bm}
\jvol{XX}
\jnum{XX}
\jmonth{XXXXX}
\paper{1234567}
\pubyear{2022}
\doiinfo{TAES.2022.Doi Number}

\newcommand{\Adjx}{\boldsymbol{A}\mathrm{d}_{\bchi}}
\newcommand{\bchi}{\boldsymbol{\chi}}
\begin{document}

\title{Semi-Aerodynamic Model Aided Invariant Kalman Filtering for UAV Full-State Estimation}

\author{Xiaoyu Ye}

\author{Fujun Song}

\author{Zongyu Zhang}

\author{Rui Zhang}

\author{Qinghua Zeng}
\affil{School of Aeronautics and Astronautics, Sun Yat-sen University, Shenzhen, China} 


\receiveddate{Manuscript received XXXXX 00, 0000; revised XXXXX 00, 0000; accepted XXXXX 00, 0000.\\
This research was funded by the National Natural Science Foundation of China, grant number 61174120.}

\authoraddress{The authors are with the School of Aeronautics and Astronautics, Sun Yat-sen University, Shenzhen 518107, China.
(e-mail: \href{mailto:yexy39@mail2.sysu.edu.cn}{yexy39@mail2.sysu.edu.cn}, \href{mailto:zqinghua@mail.sysu.edu.cn}{zqinghua@mail.sysu.edu.cn}, \href{mailto:songfj3@mail2.sysu.edu.cn}{songfj3@mail2.sysu.edu.cn}, \href{mailto:zhangzy259@mail2.sysu.edu.cn}{zhangzy259@mail2.sysu.edu.cn}, \href{mailto:zhangr365@mail2.sysu.edu.cn}{zhangr365@mail2.sysu.edu.cn}).}

\corresp{ {\itshape (Corresponding author: Qinghua Zeng)}.}

\supplementary{The opensource code is then uploaded to github.}

\markboth{AUTHOR ET AL.}{SHORT ARTICLE TITLE}
\maketitle

\begin{abstract}    
Due to the state trajectory-independent features of invariant Kalman filtering (InEKF), it has attracted widespread attention in the research community for its significantly improved state estimation accuracy and convergence under disturbance. In this paper, we formulate the full-source data fusion navigation problem for fixed-wing unmanned aerial vehicle (UAV) within a framework based on error state right-invariant extended Kalman filtering (ES-RIEKF) on Lie groups. We merge measurements from a multi-rate onboard sensor network on UAVs to achieve real-time estimation of pose, air flow angles, and wind speed. Detailed derivations are provided, and the algorithm's convergence and accuracy improvements over established methods like Error State EKF (ES-EKF) and Nonlinear Complementary Filter (NCF) are demonstrated using real-flight data from UAVs. Additionally, we introduce a semi-aerodynamic model fusion framework that relies solely on ground-measurable parameters. We design and train an Long Short Term Memory (LSTM) deep network to achieve drift-free prediction of the UAV's angle of attack (AOA) and side-slip angle (SA) using easily obtainable onboard data like control surface deflections, thereby significantly reducing dependency on GNSS or complicated aerodynamic model parameters. Further, we validate the algorithm's robust advantages under GNSS denied, where flight data shows that the maximum positioning error stays within 30 meters over a 130-second denial period. To the best of our knowledge, this study is the first to apply ES-RIEKF to full-source navigation applications for fixed-wing UAVs, aiming to provide engineering references for designers. Our implementations using MATLAB/Simulink will open source.
\end{abstract}


\section{INTRODUCTION}
Due to the flexibility and payload capability of UAVs, they now play a significant role in both civilian and military domains. Onboard navigation systems, as the sensory organs of UAVs, integrate multisource data to provide accurate and robust reliable navigation information \cite{ye2023review,bijjahalli2020advances}. The primary purpose of navigation systems is to estimate the current state of the UAVs using a limited set of measurements available at current given time. These states include pose and velocity of the aircraft, flight conditions closely related to aerodynamic angles such as AOA and SA, as well as the wind disturbances affecting during flight \cite{yang2022variational}. A comprehensive and robust navigation system needs to fully utilize available onboard information to enhance the estimation capabilities for the above-mentioned states. However, these information are often heterogeneous, multi-rate, multi-dimensional, and has varied error characteristics, which poses challenges for the holistic design of navigation systems \cite{meng2022resilient}.

The widespread development of navigation algorithms is closely tied to Kalman filter (KF) techniques \cite{nazarahari202140,yan2020}. When both the system process model and measurement model are linear, and the noise is white noise (Gaussian noise), KF provides an optimal method for state estimation, where "optimal" refers to the minimum mean squared error \cite{kalman1960,yanbook}.

Although KF offers an optimal method for linear filtering, the dynamics of most actual UAV systems are nonlinear, and many sensor models are also commonly defined as nonlinear. To address this, the Extended Kalman Filter (EKF) effectively resolves the inability of the KF to handle nonlinearities. It uses a Taylor series expansion to linearize the system process and measurement models around the current state estimates, ignoring higher-order terms \cite{bailey2006consistency}, thereby transforming them into first-order linear models. Due to its low computational complexity and accurate performance, EKF has quickly been widely applied in the field of robotic state estimation. Christophersen et al. \cite{christophersen2006compact} used a 16-dimensional EKF to fuse GPS and INS data, while Bristeau et al. \cite{bristeau2010hardware} used a 23-dimensional EKF to merge IMU, GPS, magnetometer, and barometer data. In the PX4 ECL (Estimation and Control Library), which aims for open-source engineering implementation of UAVs, a std-EKF (standard extended Kalman filter) framework is used to estimate the state of the UAV, fusing data from redundant onboard sensors \cite{ECL}. However, it should be noted that if the system state estimation is poorly initialized or if the nonlinearity is too strong at any given moment, the filter is prone to divergence. Additionally, because EKF linearizes the estimated system at the current moment, in certain scenarios, some unobservable states may falsely appear to be observable after linearization, leading to inconsistency \cite{huang2013improving}, which negatively impacts the estimation system.

To mitigate the impact of linearization, the ES-EKF, which uses state errors such as pose error and IMU bias error as system states, demonstrates superior performance and better complies with linearization requirements compared to the standard EKF (std-EKF), especially under small angle noise assumptions \cite{Hartley2020}. Mourikis et al. \cite{mourikis2007multi} proposed the Multi-State Constraint Kalman Filter (MSCKF), applying the EKF algorithm to vision-aided inertial systems, forming the foundation for many filter-based Visual-Inertial Simultaneous Localization and Mapping (VI-SLAM) systems. Li et al. \cite{li2012improving} further optimized MSCKF, improving the consistency and accuracy of algorithm estimates. Huang et al. \cite{geneva2020openvins} introduced the OpenVINS open-source framework, employing First-Estimates Jacobians (FEJ) to enhance estimate consistency. Under small angle error assumptions, error dynamics can be better linearized compared to system dynamics, better adhering to the linearity assumption. However, it should be noted that while the error EKF framework can better handle system errors, its current estimates still depend on the current system state variables. If there is a deviation during system initialization or if the estimate strays, it may lead to system state estimate divergence, thereby reducing the filter's performance and consistency \cite{madyastha2011extended}.

In recent years, Lie group theory, which describes 3D poses, has received widespread attention in the field of pose estimation and SLAM \cite{qin2018vins,xiwei2022factor,dellaert2021factor}. The use of symmetry and Lie groups in observer design is also increasingly acknowledged and is playing an ever-larger role. Lie groups provide a natural and continuous framework for describing the three-dimensional rotation and transformation of a body. Traditional standard KFs establish systems based on Cartesian coordinates, but this representation method may overlook certain nonlinear characteristics, especially in the field of pose estimation in three-dimensional space. Many publications point out that poses are better established on Lie group-based manifolds, as they offer rigorous calculus on the nonlinear spaces of rotation and motion, and properly address pose uncertainties, optimization steps, and increments. 

Combining preservation symmetry theory with error state EKF (Extended Kalman Filter) has led to the development of the invariant EKF (InEKF). Barrau et al. \cite{Barrau2017TAC,barrau2018invariant} have expanded on the existing theory of invariant Kalman filters and demonstrated their convergence. A significant advantage of InEKF is that it constructs system states based on manifold theory. By exploiting the symmetry features of this theory and ensuring that the dynamics satisfy the group-affine properties, the error is constrained within logarithmic linear differential equations on Lie algebra. This property implies that the transition process of error system dynamics is independent of the current state, ensuring that even if the initial state is inaccurate, the linearization's accuracy will not be affected. Therefore, compared to standard EKFs based on error form, InEKF exhibits superior accuracy and consistency.

InEKF research has gained widespread attention and development in recent 5 years. Zhang et al. \cite{Zhang2017} applied RI-EKF to the field of SLAM and conducted a deep analysis on its convergence and consistency. On the other hand, Hartley et al. \cite{Hartley2020} innovatively designed an InEKF with corrections for contact kinematics, the experimental results of which demonstrated the method's performance advantages. Concurrently, Cui et al. \cite{Cui2021} derived an LG-EKF (Lie Group EKF) for a MIMU/GNSS/magnetometer integrated navigation system based on Lie groups. Similarly, Potokar et al. \cite{Potokar2021} applied RIEKF in the underwater robotics, successfully extending Doppler Velocimeter (DVL) and depth measurements into an invariant framework. In the field of visual SLAM, Liu et al. \cite{Changwu2022} proposed InGVIO, a tightly-coupled GNSS/visual/IMU system that has been successfully validated on fixed-wing platforms. Yang et al. \cite{yang2022decoupled} implemented RIEKF in visual-inertial fusion, utilizing invariant state representation to maintain system consistency. In the initial alignment field, a study \cite{Du2022} introduced an invariant UKF fusing GNSS with MIMU to improve navigation performance under large initial alignment errors. Considering that different sensors usually operate on different symmetry structures due to discrepancies in coordinate systems, Hwang et al. \cite{Hwang2022} proposed a federated InEKF structure where sub-filters LI-EKF and RI-EKF are updated independently, thus ensuring that the measurement update process is trajectory independent.

Although InEKF and Lie group theory have significant applications in multiple domains, challenges still exist in the flight state estimation for fixed-wing UAVs, particularly in complex state estimation issues like angle of airflow and wind disturbances. The primary reason lies in the variability of airflow leading to aerodynamic uncertainties, which in turn impacts the effectiveness of state estimation. Currently, the field of model-aided state estimation is gradually diverging into two paths: (1) relying on accurate aerodynamic models for corrections, and (2) depending solely on kinematic model constraints. Specifically, Tian et al. \cite{tian2018design} utilized EKF and CF filters aided by aerodynamic models to achieve AOA/SA estimation independent of direct GNSS airflow angle measurements, ensuring estimation performance even under high maneuverability, although wind effects were not considered. Youn et al. \cite{youn2020aerodynamic} also used aerodynamic models and accounted for wind disturbances, utilizing aerodynamic coefficients and control signals to predict the body's angular velocity, thereby improving attitude estimation accuracy, even maintaining stable attitude estimation in the absence of angular velocity signals. Tian et al. \cite{tian2021wind} conducted a review on wind modeling and estimation methods, pointing out that the choice of wind measurement and estimation methods is primarily application-driven. For GNC development, the preference is towards using conventional sensors for filter-based estimation, and most wind field modeling adopts a first-order random walk model. Given that small UAVs are usually low-cost and obtaining their accurate aerodynamic models is challenging, researchers are increasingly focusing on model-independent methods to enhance generalizability and reduce experimental costs. Yang et al. \cite{yang2022model} proposed a model-free distributed state estimation algorithm that not only avoids reliance on aerodynamic parameters but also successfully integrates multi-source data to estimate wind and airspeed under wind disturbances. However, the std-EKF used in this method may have limitations in accuracy in certain highly nonlinear scenarios.

To the best of our knowledge, no existing work has established on full-source state estimation for fixed-wing UAVs based on InEKF. Given that the state estimation of fixed-wing UAVs involves multiple coordinate systems and is closely related to wind disturbances, the estimation accuracy of rotations between these coordinate systems becomes a critical factor leading to discrepancies in pose accuracy. Particularly during initial pose offsets, the rapid convergence of the algorithm becomes a key evaluation metric. Previous works \cite{Potokar2021,Hartley2020,Zhu2022} have explored the convergence characteristics of different algorithms during initial value deviations through simulations. To strengthen the persuasiveness of these results, we used real flight data for parameter bias simulations. Additionally, we take full advantage of the resources of onboard sensors and propose a semi-dynamic model predictive compensation method. This method relies only on real flight data and some measurable UAV parameters such as size, weight, wing area, and chord length. We extend traditional full-source sensor measurements to constraints that depend on aerodynamic models, such as airspeed and control surface deflections, aiming to enhance estimation accuracy, especially for resilient navigation requirements under GNSS denied. Furthermore, we propose using LSTM deep networks to achieve drift-free predictions of AOA and SA without relying on GNSS measurement information. LSTM networks have great potential for predicting sequential data and offer advantages over least squares, BP neural networks, and the like. We also embed the LSTM network architecture trained with trial flight data into the full-source invariant Kalman framework.

\begin{figure}[tbp]
\centerline{
    \includegraphics[width=18.5pc]{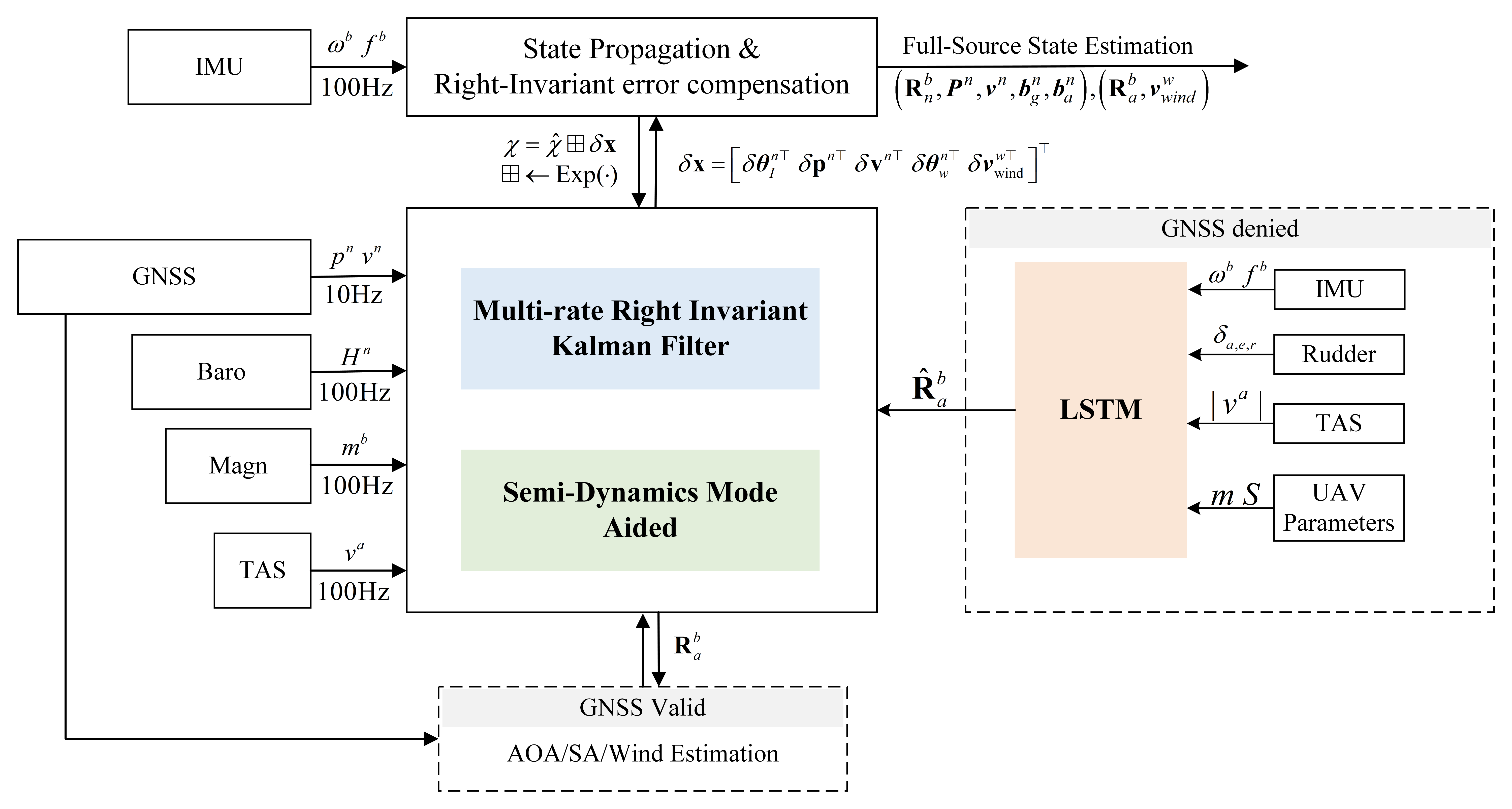}}
    \centering
    \caption{The framework for the ES-RIEKF algorithm in fixed-wing UAV.}
    \label{algorithm structure}
\end{figure}

In summary, this paper focuses on the full-source state estimation of fixed-wing UAVs based on invariant Kalman filtering with semi-aerodynamic model assistance. The main algorithmic framework of the paper is shown in Fig. \ref{algorithm structure}. The main contributions of the paper are as follows:
\begin{enumerate}[(1)]
\item We propose a multi-rate full-source navigation algorithm framework for fixed-wing UAVs using ES-RIEKF, integrating onboard sensor networks including IMU, GNSS, magnetometer, barometer, airspeed meter, and control surface deflections. This enhances the convergence and accuracy of the algorithm under disturbances.
\item Accurate aerodynamic model parameters are needed for predicting airflow angles in low-cost UAVs. Addressing the difficulty of acquiring UAV aerodynamic model parameters, we propose a semi-model fusion framework that relies only on ground-measurable parameters. Using LSTM networks, we incorporate easily obtainable sensor data like control surface deflections to make sequence-based predictions for AOA/SA airflow angles and integrate this network into the ES-RIEKF framework.
\item Through offline validation with flight data, ES-RIEKF shows advantages in pose estimation accuracy and convergence rate compared to ES-EKF, ES-LIEKF, and NCF. As a full-source navigation algorithm, it performs excellently in scenarios where GNSS signals are denied.
\end{enumerate}

This work addresses significant challenges in fixed-wing UAV state estimation, particularly in the presence of complex aerodynamics and wind disturbances. The proposed algorithms and frameworks pave the way for more robust and reliable fixed-wing UAV navigation, especially in GNSS denied environments.

The rest of this article is organized as follows. Section \ref{section2} primarily introduces the theoretical foundation related to Lie groups and invariant Kalman filtering. Section \ref{section3} focuses on designing the filter structure, and deriving the left and right invariant error dynamics equations along with the full-source sensor measurement update equations. Section \ref{section4} mainly conducts experimental validation of the algorithm from four dimensions, and compares and evaluates the algorithm. Section \ref{section5} concludes and provides future perspectives. Section \ref{appendix} contains the derivation of key formulas.

\section{InEKF theory}\label{section2}
In this section, we provide a brief introduction to the essential Lie group theory required and use it in Section \ref{section3} to succinctly explain and derive the RInEKF (and LInEKF) along with some of its key properties.

\subsection{Lie Group and Lie Algebras}
A Lie group is both a group and a smooth manifold. A Lie group matrix \( G \) is a subset of square invertible \( N \times N \) matrices \( \mathscr{M}_N(\mathbb{R}) \), and satisfies the following properties:
\[
\boldsymbol{I}_N \in G, \quad \forall \bchi \in \boldsymbol{G}, \bchi^{-1} \in \boldsymbol{G}, \quad \forall \bchi_1, \bchi_2 \in \boldsymbol{G}, \bchi_1 \bchi_2 \in \boldsymbol{G}
\]
here, \( \boldsymbol{I}_N \) is the \( N \times N \) identity matrix for \( \mathbb{R}^N \). The subset \( \boldsymbol{G} \) is not a vector space but can be considered as a curved manifold space. At each spatial point \( \bchi \in \boldsymbol{G} \), we can express the manifold's \( \boldsymbol{T}_{\bchi} \boldsymbol{G} \) as a tangent plane, referred to as the tangent space at point \( \bchi \). It is written as \( \frac{d}{d t} \boldsymbol{\gamma}(0) \), where \( \gamma: \mathbb{R} \rightarrow \boldsymbol{G} \) is a smooth curve in \( \boldsymbol{G} \) and satisfies \( \boldsymbol{\gamma}(0)=\bchi \). The elements in this space are called tangent vectors.

The unit vector \(\boldsymbol{I}_N\) at the tangent space \(\boldsymbol{T}_{I_N} \boldsymbol{G}\) is known as the Lie algebra, which plays an extremely crucial role. The space of Lie algebra is isomorphic to Cartesian space, thus elements of the tangent space can be transformed both into Cartesian space and manifold space. The Lie algebra is denoted as \(\mathfrak{g}\), and its dimension \(d\) defines the dimension of the group \(\boldsymbol{G}\) itself. For \(\boldsymbol{\xi} \in \mathbb{R}^d\), there is always an invertible mapping \(\mathbb{R}^d \rightarrow \mathfrak{g}\) from \(\mathfrak{g}\) to \(\mathbb{R}^d\). We take \( \boldsymbol{\xi}^{\wedge} \in \mathfrak{g}\) as the corresponding element in Lie algebra \(\mathfrak{g}\), and consider the mapping \(\boldsymbol{\xi} \mapsto \boldsymbol{\xi}^{\wedge}\) as a linear transformation. The operations \((\cdot)^{\wedge}\) and \((\cdot)^{\vee}\) denote the mapping and its inverse between Euclidean vector space \(\mathbb{R}\) and the Lie algebra \(\mathfrak{g}\), as expressed in the following equation:
\begin{equation}
    (\cdot)^{\wedge}: \mathbb{R}^{\operatorname{dim} \mathfrak{g}} \rightarrow \mathfrak{g}, \quad(\cdot)^{\vee}: \mathfrak{g} \rightarrow \mathbb{R}^{\operatorname{dim} \mathfrak{g}}
\end{equation}
\subsection{Exponential and Logarithmic Mapping}
The exponential mapping, denoted as \(\exp{(\cdot)}\), precisely transforms elements from the Lie algebra to the Lie group, an operation known as retraction. Conversely, the \(\log(\cdot{})\) operation unfolds or projects elements from the Lie group back to the Lie algebra, as shown in the following equations:
\begin{equation}
\text{Exp}(\cdot) \rightarrow \exp \left((\cdot)^{\wedge}\right): \mathbb{R}^{\operatorname{dim} \mathfrak{g}} \rightarrow \boldsymbol{G},
\end{equation}
\begin{equation}
\text{Log}(\cdot) \rightarrow \log \left( (\cdot)^{\vee} \right): \boldsymbol{G} \rightarrow \mathbb{R}^{\operatorname{dim} \mathfrak{g}}
\end{equation}
\subsection{Addition and Subtraction Operators }
By defining addition and subtraction operators on the manifold, we can introduce incremental elements on manifold, thereby controlling the movement of the manifold in Lie group space. Due to the non-commutative nature of the composition, the operations are categorized into left-multiplication and right-multiplication based on the position of the incremental element $\tau $, as shown in the \eqref{plus/add op}:
\begin{equation}\label{plus/add op}
\begin{array}{cc}
   \mathcal{Y}=\bchi \boxplus{ }^{\bchi} \boldsymbol{\tau} \triangleq \bchi \cdot \operatorname{Exp}\left({ }^{\bchi} \boldsymbol{\tau}\right) \in \mathcal{M}:  &  \text{left}-\boxplus\\
    \mathcal{Y}={ }^{\mathcal{E}} \boldsymbol{\tau} \boxplus \bchi \triangleq \operatorname{Exp}\left({ }^{\mathcal{E}} \boldsymbol{\tau}\right) \cdot \bchi \in \mathcal{M}: &  \text{right}-\boxplus\\
\end{array}
\end{equation}

\subsection{Adjoint Operator}
To transform elements between the navigation coordinate system and the body coordinate system, the adjoint matrix is needed. If we make the left and right operations in \eqref{plus/add op} equal, that is, \( \bchi \cdot \operatorname{Exp}({ }^{\bchi} \boldsymbol{\tau}) = \operatorname{Exp}({ }^{\mathcal{E}} \boldsymbol{\tau}) \cdot \bchi \), this establishes the relationship between local and global tangent elements. The derivation is as follows:
\begin{equation}
\begin{aligned}
\operatorname{Exp}\left({ }^{\mathcal{E}} \boldsymbol{\tau}\right) \bchi & =\bchi \operatorname{Exp}\left({ }^{\bchi} \boldsymbol{\tau}\right) \\
\exp \left({ }^{\mathcal{E}} \boldsymbol{\tau}^{\wedge}\right) & =\bchi \exp \left({ }^{\bchi} \boldsymbol{\tau}^{\wedge}\right) \bchi^{-1}=\exp \left(\bchi^{\bchi} \boldsymbol{\tau}^{\wedge} \bchi^{-1}\right) \\
^{\mathcal{E}}{\boldsymbol{\tau}^{\wedge}} & =\bchi^{\bchi} \boldsymbol{\tau}^{\wedge} \bchi^{-1}
\end{aligned}
\end{equation}

The \textbf{adjoint} \(\Adjx\) is defined as a mapping:
\begin{equation}\label{adjoint}
    \operatorname{Ad}_{\bchi}: \mathfrak{g} \rightarrow \mathfrak{g}, \quad \tau^{\wedge} \mapsto \Adjx \left(\tau^{\wedge}\right) \triangleq \bchi \tau^{\wedge} \bchi^{-1},
\end{equation}
Since the adjoint mapping is linear, we can express the mapping of the adjoint matrix as \( \Adjx \left(\chi^{\wedge}\right)=\left(\Adjx \boldsymbol{\xi}\right)^{\wedge} \).

\subsection{Group Affine}
If the IEKF system state equation $f( \cdot )$ meets the following conditions, it is referred to as a group-affine system.

\begin{thm}\label{thm1}
The following equivalences hold:
\begin{align}
&\forall \bchi, \boldsymbol{v} \in \boldsymbol{G}, \forall \boldsymbol{u} \quad f(\bchi \boldsymbol{v}, \boldsymbol{u}) = f(\bchi, \boldsymbol{u}) f\left(\boldsymbol{I}_N, \boldsymbol{u}\right)^{-1} f(\boldsymbol{v}, \boldsymbol{u}) \label{group offline}\\
&f \, \text{satisfies} \eqref{group offline} \Leftrightarrow \text{there exists a mapping } g(\cdot), \nonumber \\
&\forall \bchi, \boldsymbol{v}, \boldsymbol{u}, f(\boldsymbol{v}, \boldsymbol{u})^{-1} f(\bchi, \boldsymbol{u}) = g\left(\boldsymbol{v}^{-1} \bchi, \boldsymbol{u}\right) \nonumber
\end{align}
\end{thm}
For any \( \boldsymbol{u} \in \mathbb{R}^{d \times d} \), there exists \( \boldsymbol{F} \in \mathbb{R}^{d \times d} \) such that \( \forall \boldsymbol{\xi} \in \mathbb{R}^d, g(\exp (\boldsymbol{\xi}), \boldsymbol{u}) \triangleq \exp (F \boldsymbol{\xi}) \). Therefore, the function \( g(\cdot) \) can be expressed in terms of the simple matrix \( \boldsymbol{F} \).

\subsection{InEKF}
For a detailed theoretical analysis of InEKF, see  \cite{Barrau2017TAC}, \cite{barrau2018invariant}, and \cite{Hartley2020}. The main difference between InEKF and traditional EKF lies in the different forms of error definitions. The state update process at time \( t \) and state \( \chi_t \) can be described as follows:
\begin{equation}\label{InEKF1}
    \bchi_{n|n-1} = f\left( \bchi_{n-1|n-1} , \boldsymbol{u}_n \right) + \bchi \boldsymbol{w}^{\wedge} 
\end{equation}
In (\ref{InEKF1}), \( \hat{\bchi} \in \boldsymbol{G} \) represents the estimated system state, and \( \boldsymbol{u}_n \) is the known system input, such as the output from gyroscopes and accelerometers. \( \boldsymbol{w}^{\wedge} \in \mathfrak{g} \), where \( \boldsymbol{w} \in \mathbb{R}^d \), represents Gaussian white noise. The hat notation \( \hat{(\cdot)} \) is used to indicate estimated values in the system. For the theoretical value \( \bchi \) and the estimated value \( \hat{\bchi} \), the error $\boldsymbol{\eta}$ is represented in the following left-invariant/right-invariant form:
\begin{equation}
    \begin{array}{cc}
        \boldsymbol{\eta}^L = \boldsymbol{\chi}^{-1} \hat{\bchi} = \operatorname{Exp}(\boldsymbol{\xi}^{L}) &  (\text{left invariant error})\\
        \boldsymbol{\eta}^R =  \hat{\bchi} \bchi^{-1} = \operatorname{Exp}(\boldsymbol{\xi}^{R}) & (\text{right invariant error})
    \end{array}
\end{equation}
where \( \boldsymbol{\xi}^{L} \) and \( \boldsymbol{\xi}^{R} \) are linear invariant errors. If the error system satisfies group affinity\ref{thm1}, then the invariant error differential equation can be expressed as follows:
\begin{equation}
\begin{aligned}
& \frac{d}{d t} \boldsymbol{\eta}^{L}=g_{u}^{L}\left(\boldsymbol{\eta}^{L}\right)-\left(\boldsymbol{w}^{\wedge}\right) \boldsymbol{\eta}^{L} \\
& \frac{d}{d t} \boldsymbol{\eta}^{R}=g_{u}^{R}\left(\boldsymbol{\eta}^{R}\right)-\left(\operatorname{Ad}_{\hat{\bchi}} \boldsymbol{w}^{\wedge}\right) \boldsymbol{\eta}^{R}
\end{aligned}
\end{equation}
where
\begin{equation}
\begin{aligned}
& g_{u}^{L}\left(\boldsymbol{\eta}^{L}\right)=f_{u}\left(\boldsymbol{\eta}^{L}\right)-f_{u}(\bm{I}) \boldsymbol{\eta}^{L} \\
& g_{u}^{R}\left(\boldsymbol{\eta}^{R}\right)=f_{u}\left(\boldsymbol{\eta}^{R}\right)-\boldsymbol{\eta}^{R} f_{u}(\bm{I}) 
\end{aligned}
\end{equation}
Then, based on Theorem \ref{thm1}, an error update equation can be established:
\begin{equation}
    \frac{d \boldsymbol{\xi}}{dt} = \mathbf{F} \boldsymbol{\xi} + \mathbf{G} \boldsymbol{w} 
\end{equation}

For InEKF measurement updates, the observable quantities for left-invariant and right-invariant Kalman filters can be described using the following formula:
\begin{equation}
\begin{array}{cc}
    \mathbf{Z}^L = \mathbf{X} \mathbf{b} + W & \text{(left invariant)}
    \end{array}
\end{equation}
\begin{equation}\label{IEKF Mea update}
\begin{array}{cc}
    \mathbf{Z}^R = \mathbf{X}^{-1} \mathbf{b} + W & \text{(right invariant)} \\
\end{array}
\end{equation}

Among them, \( \mathbf{b} \) is a known coefficient matrix, and \( W \) is zero-mean white noise \( w \sim \mathcal{N}(0, \sigma^2) \). The innovation for the InEKF can be calculated as follows:
\begin{equation}
\begin{array}{cc}
    \bm{\gamma}^L = \hat{\mathbf{X}}^{-1} \left( \mathbf{Z}^L - \hat{\mathbf{Z}}^L \right) & \text{(left invariant)} \\
\end{array}
\end{equation}
\begin{equation}
\begin{array}{cc}
    \bm{\gamma}^R = \hat{\mathbf{X}} \left( \mathbf{Z}^R - \hat{\mathbf{Z}}^R \right) & \text{(right invariant)} \\
\end{array}
\end{equation}
where, \( \hat{\mathbf{Z}} \) represents the current state estimate. According to \( \bm{\eta} = \exp{(\boldsymbol{\xi})} \approx I + \boldsymbol{\xi}^{\wedge} \) and \( \bm{\eta}^R = \hat{\mathbf{X}} \mathbf{X}^{-1} \), we can obtain:
\begin{equation}
    \begin{aligned}
        \bm{\gamma}^R &= \hat{\mathbf{X}} \left( \mathbf{Z}^R - \hat{\mathbf{Z}}^R \right) = \hat{\mathbf{X}} \left(\mathbf{X}^{-1} \mathbf{b}+ W - \hat{\mathbf{X}}^{-1} \mathbf{b} \right) \\
        &= \hat{\mathbf{X}} \mathbf{Z}^R - \mathbf{b}
        \approx \bm{\xi}^{R\wedge} \mathbf{b} + \hat{\mathbf{X}} W \triangleq - \mathbf{H}^R \bm{\xi}^R + \hat{\mathbf{X}} W
    \end{aligned}
\end{equation}
A similar conclusion can be drawn for the left-invariant error \( \bm{\xi}^L \):
\begin{equation}
    \begin{aligned}
        \bm{\gamma}^L &= \hat{\mathbf{X}}^{-1} \left( \mathbf{Z}^L - \hat{\mathbf{Z}}^L \right) = \hat{\mathbf{X}}^{-1} \left(\mathbf{X} \mathbf{b}+ W - \hat{\mathbf{X}} \mathbf{b} \right) \\
        &= \hat{\mathbf{X}}^{-1} \mathbf{Z}^L - \mathbf{b} 
        \approx \bm{\xi}^{L\wedge} \mathbf{b} + \hat{\mathbf{X}}^{-1} W \triangleq - \mathbf{H}^L \bm{\xi}^L + \hat{\mathbf{X}}^{-1} W
    \end{aligned}
\end{equation}
At this point, the observation matrices \( \mathbf{H}^L \) and \( \mathbf{H}^R \) for the left-invariant and right-invariant errors \( \bm{\xi}^L \) and \( \bm{\xi}^R \) can be calculated.

\section{Filter Design}\label{section3}
In this section, we primarily focus on deriving the update equations for multi-sensor fusion in both left-invariant and right-invariant Kalman filter. This includes the prediction models for left and right invariants as well as the measurement update models for sensors from all sources.

\subsection{State}
\begin{figure}[htbp]
\centerline{
    \includegraphics[width=1\linewidth]{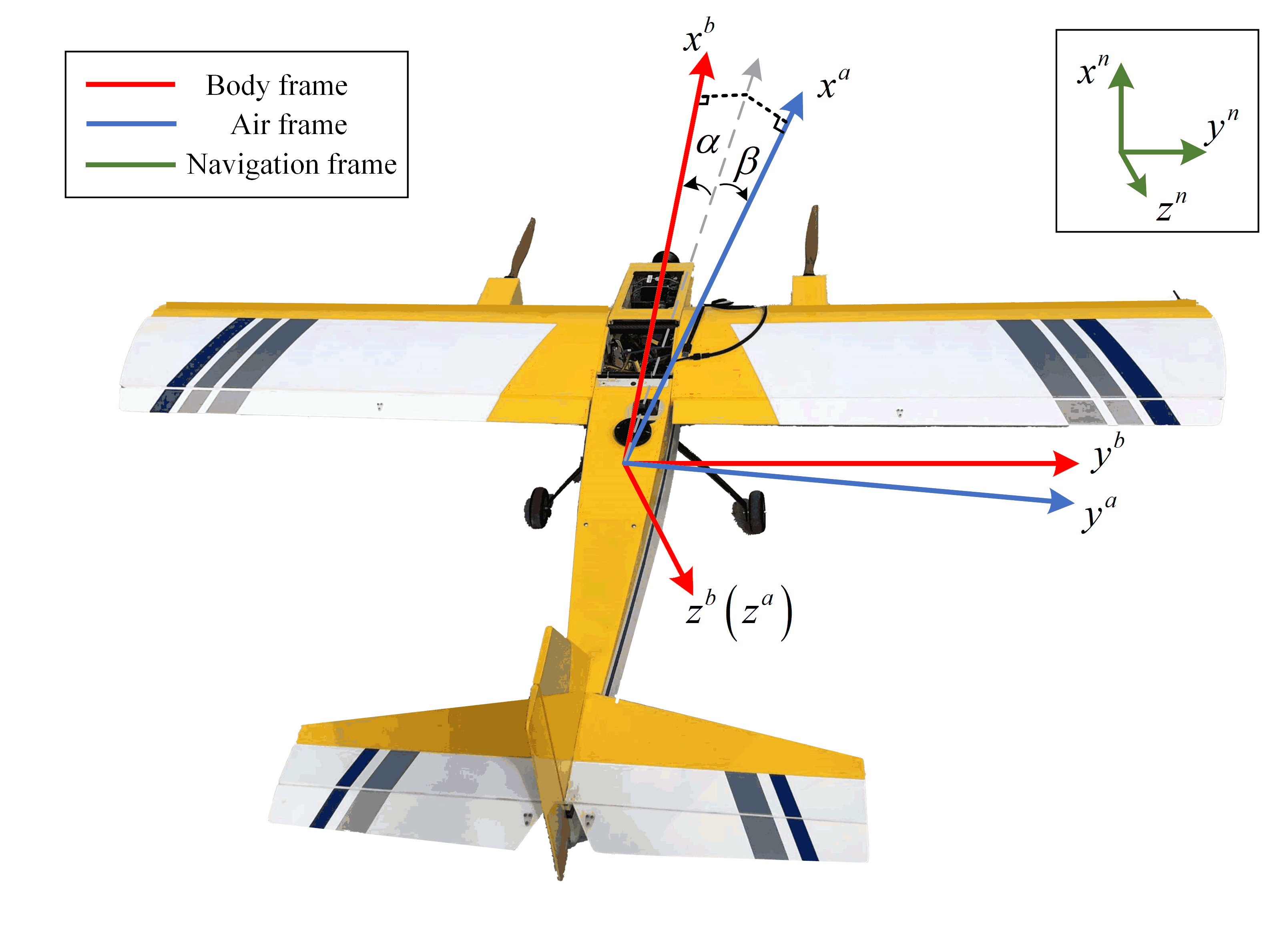}}
    \centering
    \caption{UAV frame definition.}
    \label{UAV frame}
\end{figure}

The coordinate systems used in this paper are defined as follows: Earth-Centered-Earth-Fixed (ECEF) coordinate system, represented by \{e\}. North-East-Up (ENU) coordinate system, also known as the navigation coordinate system, represented by \{n\}. IMU coordinate system, with its axes oriented in the front-right-down direction, represented by \{b\}. Airflow coordinate system, as shown in the Fig.~\ref{UAV frame}, represented by \{a\}. The symbol \(\hat{\cdot}\) indicates estimated values, and \(\delta\) represents state error variables.
The definitions of formula variables in this paper are as follows: The superscript on the variable indicates the corresponding coordinate system, and the subscript indicates the corresponding state. Specifically, \( \mathbf{R}_{\text{frame1}}^{\text{frame2}} \in \mathbb{S O}(3) \) represents the attitude rotation matrix from frame1 to frame2.

The definition of the full system state variables is as follows: 
\begin{equation}
    \bm{\chi}=\left( \bm{\chi}_{IMU}, \bm{\chi}_{ADS} \right)
\end{equation}
In this definition, the IMU state is represented as \( \bm{\chi}_{\text{IMU}} = \left( \mathbf{R}_b^n, \mathbf{v}^n, \mathbf{p}^n, \mathbf{b}_g^n, \mathbf{b}_a^n \right) \), where \( \left( \mathbf{R}_b^n, \mathbf{v}^n, \mathbf{p}^n \right) \in \mathbb{S E}_2(3) \) indicates the pose and velocity of the vehicle. \( \left( \mathbf{b}_g^n, \mathbf{b}_a^n \right) \in (\mathbb{R}^3)^2 \) represents the tri-axial bias error for the IMU's gyroscope and accelerometer. The Air Data System (ADS) state \( \bm{\chi}_{\text{ADS}} = \left( \mathbf{R}_a^b, \mathbf{v}_{\text{wind}}^n \right) \), where \( \mathbf{R}_a^b \in \mathbb{S O}(3) \) is the attitude rotation matrix of the air flow coordinate system relative to the body frame, and \( \mathbf{v}_{\text{wind}}^n \in \mathbb{R}^3 \) indicates the wind speed magnitude.

The right (left) invariant error can be defined as:
\begin{equation}
\begin{aligned}
\delta \mathbf{x}&= (\delta \bchi_{I},\delta \mathbf{x}_{b},
\delta \bchi_{A}, \delta \mathbf{v}^n_{wind})\\
&= \left[
\setlength{\arraycolsep}{2pt}
\begin{array}{cccccccccc}
\delta \boldsymbol{\theta}_I^{n \top} & 
\delta \mathbf{v}^{n \top} & 
\delta \mathbf{p}^{n \top}  & | &
\delta \bm{b}_g^{b \top} & 
\delta \bm{b}_a^{b \top} & | &
\delta \boldsymbol{\theta}_a^{b \top} &| & 
\delta \mathbf{v}^{n \top}_{wind}
\end{array}\right]^{\top}
\end{aligned}
\end{equation}
The error form in RInEKF satisfies the following \eqref{RInEKF def}:
\begin{equation} \label{RInEKF def}
\begin{aligned}
\bm{\chi} &= \hat{\bm{\chi}} \boxplus \delta \mathbf{x}\\
&=\left( \exp \left(\delta \bm{\chi}_{I}\right) \hat{\mathbf{x}}_{I_R},
\hat{\mathbf{x}}_b+\delta \mathbf{x}_b,
\exp \left(\delta \mathbf{\chi}_{A}\right) \hat{\mathbf{x}}_{A},
\hat{\mathbf{x}}_{vw}+\delta \mathbf{x}_{vw}
\right) \\
 &= \left[\begin{array}{c}
   \exp(\delta \boldsymbol{\theta}_I) \mathbf{R}_b^n \\
   \exp(\delta \boldsymbol{\theta}_I) \mathbf{\hat{p}}^n +\mathbf{J}_l(\delta \boldsymbol{\theta}_I) \delta \mathbf{p} \\
   \exp(\delta \boldsymbol{\theta}_I) \mathbf{\hat{v}}^n  + \mathbf{J}_l(\delta \boldsymbol{\theta}_I)\delta \mathbf{v} \\
   \mathbf{\Hat{b}}_g + \delta \mathbf{b}_g \\
   \mathbf{\Hat{b}}_a + \delta \mathbf{b}_a \\
   \exp(\delta \boldsymbol{\theta}_a) \mathbf{R}_a^b \\
   \hat{\mathbf{v}}^n_{w} + \delta \mathbf{v}^n_{w}\\
\end{array}\right]
\end{aligned}
\end{equation}
where,
\begin{equation}
    \boldsymbol{\chi}_{I_R}=\left[\begin{array}{c|c}
\mathbf{R}_b^n & \mathbf{v}^n \,\,\,\mathbf{p}^n \\
\hline \mathbf{0}_{2\times 3} & \mathbf{I}_2
\end{array}\right] \in \mathbb{S E}_2(3)
\end{equation}
The  \(\bm{\chi}_{I_R}\) represents the right-invariant state of the system.

\subsection{Error State Propagation}
Firstly, the models for the gyroscope and accelerometer are established as follows:
The Inertial Measurement Unit (IMU) measures the angular rate \(\boldsymbol{\hat{\omega}}\) and the body acceleration \(\boldsymbol{\hat{a}}\) augmented by gravity, both in the body frame \(b\). The measurement errors include accelerometer bias \(\boldsymbol{b}_a\), gyroscope bias \(\boldsymbol{b}_g\), and additional Gaussian white noise, as shown in \eqref{IMU error model}:
\begin{equation}\label{IMU error model}
\begin{gathered}
    \boldsymbol{\hat{\omega}}^b = \boldsymbol{\omega}^b + \boldsymbol{b}_g + \boldsymbol{n}_{\omega}, \boldsymbol{n}_{\omega} \sim \mathcal{N} \left(0, \sigma_{{\omega}}^2 \right)\\
        \boldsymbol{\hat{a}}^b = \boldsymbol{a}^b + \boldsymbol{b}_a + \boldsymbol{n}_{a}, \boldsymbol{n}_{a} \sim \mathcal{N} \left(0, \sigma_{{a}}^2 \right)\\
    \end{gathered}
\end{equation}

The continuous-time state update model for the system can be represented as:
\begin{equation}
\begin{gathered}
\label{state update dynamic}
    \Dot{\mathbf{R}}_b^n = \mathbf{R}_b^n [\mathbf{\omega}^b]_{\times},\Dot{\mathbf{v}}^n = \mathbf{R}_b^n \boldsymbol{a}^b + \mathbf{g}^n,
    \Dot{\mathbf{p}}^n =  \mathbf{v}^n \\
    \Dot{\boldsymbol{b}}_g = \boldsymbol{n}_{b_g},\Dot{\boldsymbol{b}}_a = \boldsymbol{n}_{b_a}
\end{gathered}
\end{equation}
where, \( \setlength{\arraycolsep}{2pt} \mathbf{g}^n = \left[\begin{array}{ccc}
0 & 0 & -9.79 \\
\end{array} \right]^\top \) represents the projection of the gravitational acceleration in the \( n \) frame. The notation \( [\cdot]_{\times} \) signifies an anti-symmetric matrix.
The continuous-time right-invariant dynamics model of the system can be expressed as:
\begin{equation}\label{Right error continuous time}
    \delta \Dot{\mathbf{x}_R} = \mathbf{F}_t^R \delta \mathbf{x}_R + \mathbf{G}^R_t \mathbf{n}
\end{equation}
In \eqref{Right error continuous time}, \(\mathbf{F}_t^R\) represents the right-invariant state transition matrix. Similarly, \(\mathbf{G}^R_t\) represents the right-invariant noise excitation matrix.
Based on (\ref{RInEKF def}), the invariant state error portion \(\bm{\chi}_{I_R}\), which satisfies the group-affine property, can be described as follows:
\begin{equation}\label{IMU right error state definition}
\begin{aligned}
    &\delta \bm{\chi}_{I_R}=\bm{\chi}_{I_R}
    \hat{\bm{\chi}}_{I_R}^{\top}\\
    &=\left[\begin{array}{c|c}
       \mathbf{R}_b^n \hat{\mathbf{R}}_b^{n\top}  &
       \mathbf{p}^n - \mathbf{R}_b^n \hat{\mathbf{R}}_b^{n\top} \mathbf{\hat{p}}^n \,\,|\,\,
       \mathbf{v}^n - \mathbf{R}_b^n \hat{\mathbf{R}}_b^{n\top} \mathbf{\hat{v}}^n\\
       \hline
       \bm{0}_{2 \times 3} & \bm{I}_2 \\
    \end{array}\right] \\
    \end{aligned}
\end{equation}

The system's error dynamics model can be derived based on the right-invariant error definition  (\ref{IMU right error state definition}) and the system state update  (\ref{state update dynamic}). For details, see Appendix \ref{appendix B} (Note: The derivation of the left-invariant error dynamics can be found in Appendix \ref{appendix C}. For the sake of readability, this section primarily focuses on right-invariant error). Therefore, the continuous-time state transition matrix \(\mathbf{F}_t\) can be expressed as follows:
\begin{equation}\label{error F}
\begin{aligned}
    \mathbf{F}_t^R &= \left[ \begin{array}{ccc}
        \mathbf{F}_{davp} & \mathbf{F}_{davp2b} &  \boldsymbol{0}_{9\times6}\\
         \boldsymbol{0}_{9\times9}& \mathbf{I}_{9\times9} &  \boldsymbol{0}_{9\times6}\\
         \boldsymbol{0}_{6\times9}& \boldsymbol{0}_{6\times9}& \mathbf{F}_{\alpha \mathbf{v}}\\
    \end{array}\right]_{24\times24},\\
    \mathbf{G}_t^R &= \left[ \begin{array}{ccc}
       \mathbf{G}_{davp}  &  \boldsymbol{0}_{9\times9} & \boldsymbol{0}_{6\times6}\\
        \boldsymbol{0}_{9\times6} & \mathbf{I}_{9\times9} &
        \boldsymbol{0}_{9\times6}\\
         \boldsymbol{0}_{6\times6} &  \boldsymbol{0}_{6\times9}  & \mathbf{G}_{\alpha \mathbf{v}}\\
\end{array}    \right]_{24\times21}
\end{aligned}
\end{equation}

The error dynamics transition matrix \(\mathbf{F}_{davp}\), which is related to the attitude, velocity, and position, can be expressed as follows:
\begin{equation}\label{F_davp}
\begin{aligned}
    \mathbf{F}_{davp} &= \left[ \begin{array}{ccc}
         \boldsymbol{0}_{3\times3} & \boldsymbol{0}_{3\times3} & \boldsymbol{0}_{3\times3}\\
        \left[\mathbf{g}^n \right]_{\times} & \boldsymbol{0}_{3\times3} & \boldsymbol{0}_{3\times3} \\
        1/2 [\mathbf{g}^n]_{\times} & \boldsymbol{0}_{3\times3}  &\boldsymbol{0}_{3\times3} \\
    \end{array} \right]_{9\times9},\\
    \mathbf{G}_{avp} &= \left[ 
    \begin{array}{ccc}
       - \hat{\mathbf{R}}_b^n  &  \boldsymbol{0}_{3\times3}\\
        - [\mathbf{v}^n]_{\times} \hat{\mathbf{R}}_b^n & -\hat{\mathbf{R}}_b^n\\
        - [\mathbf{p}^n]_{\times} \hat{\mathbf{R}}_b^n &-\hat{\mathbf{R}}_b^n 
    \end{array}
    \right]_{9\times6}
    \end{aligned}
\end{equation}
It can be observed that \(\mathbf{F}_{davp}\) is a time-invariant constant matrix, which also reflects the invariance of error propagation.

The matrix \(\mathbf{F}_{davp2b}\) representing the relationship between the attitude velocity position (avp) error state and the bias error states \(\delta \mathbf{b}_g, \delta \mathbf{b}_a\) is shown in  (\ref{davp2b}). Due to the fact that the biases \((\mathbf{b}^n_g,\mathbf{b}^n_a) \in (\mathbb{R}^3)^2\) are \textit{state-dependent}, this is also known as \textit{imperfect InEKF}. It's worth noting that even though the Lie group does not include bias terms and does not satisfy group-affine dynamics, this InEKF still outperforms the standard EKF \cite{barrau2015non,Hartley2020}. The essence of InEKF is actually a transformation of coordinate systems. For the pose error state in the \(n\)-frame, the right-invariant pose error remains invariant. However, this invariance also extends to biases that do not satisfy the SE3 constraints. Even if the bias error is trajectory dependent, the fact that the bias exists in Cartesian coordinates makes constraining the bias very straightforward and efficient.

\begin{equation}\label{davp2b}
\mathbf{F}_{davp2b} = \left[
    \begin{array}{cc}
       - \hat{\mathbf{R}}_b^n  &  \boldsymbol{0}_{3\times3}\\
        - [\mathbf{v}^n]_{\times} \hat{\mathbf{R}}_b^n & - \hat{\mathbf{R}}_b^n \\
        - [\mathbf{p}^n]_{\times} \hat{\mathbf{R}}_b^n & - \hat{\mathbf{R}}_b^n \\
    \end{array} \right]_{9\times6}
\end{equation}

The following mainly establishes the error dynamics equation of the airframe-body rotation matrix.

AOA and SA are the angles between the velocity (also known as airflow) coordinate system $a$ and the body coordinate system $b$ on the lateral and longitudinal planes, respectively. They are essentially a form of aircraft attitude. Previous research often directly estimates the AOA and SA using the geometric relationships of velocity, and then estimates them directly using EKF. This often leads to state inconsistency because the airflow angles are closely related to the aircraft's attitude and aerodynamics. When exploring the auxiliary corrections of the dynamic model, it is essential to predict and estimate these airflow angles. However, since the iterative estimation of attitude does not meet the traditional additive rules, which is reflected in the update of the state variables and covariance, it is preferable to establish \( \alpha \) and \( \beta \) on the manifold \( \mathbf{R}_a^b \in \mathbb{SO}(3) \). The process of updating and compensating them is more suitable for using mature Lie group-related theories to improve the accuracy and consistency of estimating \( \alpha \) and \( \beta \), with the following advantages:
1) Establishing the attitude \( \mathbf{R}_a^b \) on the manifold, similar to \( \mathbf{R}_b^n \), ensures that the error state is always operating close to the true value origin, thereby ensuring that linearization effectiveness remains constant at all times.
2) Because the estimated error state is a small quantity within a unit step, this means that products of the second order and above can often be ignored, making the calculation of the Jacobians matrix very simple and fast.

Based on the kinematic relationship of the UAV in the  coordinate system $n$ and $a$, we have:

\begin{equation}\label{Rab_update}
    \mathbf{v}_{wind}^n = \mathbf{v}_{G}^n - \mathbf{R}^n_b \mathbf{R}^b_a \mathbf{v}_{TAS}^a
\end{equation}
where, 
\begin{equation} \label{Rab by alpha}
    \mathbf{R}^b_a=\left[\begin{array}{ccc}
    \cos \alpha \cos \beta & -\cos \alpha \sin \beta & -\sin \alpha \\
    \sin \beta & \cos \beta & 0 \\
    \sin \alpha \cos \beta & -\sin \alpha \sin \beta & \cos \alpha
    \end{array}\right]
\end{equation}

Based on the above equation, \( \alpha \) and \( \beta \) can be calculated as follows:
\begin{equation}\label{Rab}
    \left[\begin{array}{c}
    V^a_{TAS} \\
    \alpha \\
    \beta
    \end{array}\right]=\left[\begin{array}{c}
    \sqrt{u^2+v^2+w^2} \\
    \tan ^{-1}\left(\frac{w}{u}\right) \\
    \sin ^{-1}\left(\frac{v}{\sqrt{u^2+v^2+w^2}}\right)
    \end{array}\right]
\end{equation}
where, \( u, v, w \) are the components of the true airspeed \( V^a_{TAS} \) in the body frame. Since it is not possible to directly obtain \( \mathbf{R}^b_a \) based on (\ref{Rab_update}), solving for \( \alpha \) and \( \beta \) allows us to construct the attitude rotation matrix \( \mathbf{R}^b_a \) of the airflow frame relative to the body frame. Let \( \delta \boldsymbol{\theta}_w^{a} \) represent the estimated error of \( \mathbf{R}^b_a \), then we have:
\begin{equation}
    \mathbf{R}^b_a = \exp{(\delta \boldsymbol{\theta}_w^{a})} \mathbf{\hat{R}}^b_a
\end{equation}
based on  (\(\ref{Rab_update}\)), \(\mathbf{R}^b_a\) is calculated at each time instance and is not expressed in a time-based recursive form. It may be beneficial to represent the discrete-time transition process of the error states \(\delta \boldsymbol{\theta}^a\) and \(\delta\mathbf{v}^n_{w}\) as follows:
\begin{equation}\begin{aligned}
        \delta \boldsymbol{\theta}^a_{k+1} &= \delta \boldsymbol{\theta}^a_{k} + \mathbf{n}_{\delta \theta^a}\\
    \delta \mathbf{v}^n_{w_{k+1}} &= \delta \mathbf{v}^n_{w_{k}} + \mathbf{n}_{\delta \mathbf{v}^n_{w}}
\end{aligned}
\end{equation}

The error state update matrix for \(\mathbf{R}_a^b\) and \(\mathbf{v}_{wind}^n\), denoted as \(\mathbf{F}_{\alpha \mathbf{v}}\), is expressed as follows:
\begin{equation}
    \mathbf{F}_{\alpha \mathbf{v}} = \left[ \begin{array}{cc}
        \mathbf{I}_{3\times3} & \boldsymbol{0}_{3\times3} \\
        \boldsymbol{0}_{3\times3} & \mathbf{I}_{3\times3}\\
    \end{array}\right]_{6\times6},
    \mathbf{G}_{\alpha \mathbf{v}} = \left[ \begin{array}{cc}
         \mathbf{I}_{3\times3} & \boldsymbol{0}_{3\times3} \\
        \boldsymbol{0}_{3\times3} & \mathbf{I}_{3\times3} \\
    \end{array}\right]_{6\times6}
\end{equation}

Above all, we have established the right-invariant state transition matrix \(\mathbf{F}_t^R\). In practical calculations, it is necessary to convert the continuous-time propagation model to a discrete-time model.
\begin{equation}\label{discrete model}
\begin{gathered}
    \delta \Dot{\mathbf{x}} = \mathbf{F}_t \delta \mathbf{x} + \mathbf{G} \mathbf{n}, \delta {x}_{k+1} = \boldsymbol{\Phi} \delta \mathbf{x}_{k} + \mathbf{G} \mathbf{n}_k \\
    \dot{\boldsymbol{\Phi}}(k, k+1)=\mathbf{F} \boldsymbol{\Phi}(k, k+1), \boldsymbol{\Phi}(k, k)=\boldsymbol{I}\\
    \boldsymbol{Q}_{l+1 \mid l}=\int_{t_l}^{t_{l+1}} \boldsymbol{\Phi}\left(t_{l+1}, \tau\right) \boldsymbol{G} \boldsymbol{Q}_m \boldsymbol{G}^{\mathrm{T}} \boldsymbol{\Phi}\left(t_{l+1}, \tau\right)^{\mathrm{T}} d \tau
    \end{gathered}
\end{equation}
The noise term \(\mathbf{n}\) is given by \(\mathbf{n} = \left[
\setlength{\arraycolsep}{2pt}
\begin{array}{ccc}
     \mathbf{n}_{imu}^{\top} & \mathbf{n}_{b_m}^{\top} &\mathbf{n}_{\alpha \mathbf{v}}^{\top}\\
\end{array} \right]^{\top}\), where \(\mathbf{n}_{imu} = \left[ 
    \begin{array}{cccc}
    \mathbf{n}_a^{\top}& \mathbf{n}_g^{\top}& \mathbf{n}_{b_a}^{\top}& \mathbf{n}_{b_g}^{\top} \\
    \end{array}
    \right]^{\top}\) and \(\mathbf{n}_{\alpha \mathbf{v}} = \left[
\setlength{\arraycolsep}{2pt}
\begin{array}{cc}
     \mathbf{n}_{\delta \theta^a}^{\top} & \mathbf{v}^{n \top}_{w} \\
\end{array}
\right]^{\top}\). The descriptions of the matrices \(\mathbf{\Phi}\) and \(\mathbf{G}\) can be found in Appendix A.

\section{Multi-sensors Measurement Update}\label{section4}
Different sensors provide redundant and heterogeneous data sources for the navigation system. Due to the differences in the measurement principles of the sensors, it is necessary to model their errors separately to maximize the value of the sensor data. These are then incorporated into the InEKF framework to derive the measurement update equations. Since the IMU error model has already been described in  (\ref{IMU error model}), this section focuses on the update processes for the remaining sensors.
\subsection{GNSS Update}
When the GNSS receives positioning and velocity signals, we transfer the original measurements such as latitude, longitude, and altitude in $e$ frame to  \( n \) frame. The error equation is described as follows:

\begin{equation}\label{GNSS update}
\begin{gathered}
    \hat{\mathbf{p}}^n_G = \mathbf{p}^n_G + \boldsymbol{n}_p,\boldsymbol{n}_p \sim \mathcal{N} \left( 0,\sigma_{p_G}^2\right)\\
    \hat{\mathbf{v}}^n_G = \mathbf{v}^n_G + \boldsymbol{n}_v,\boldsymbol{n}_v \sim \mathcal{N} \left( 0,\sigma_{v_G}^2\right)
\end{gathered}
\end{equation}

According to (\ref{IEKF Mea update}), since GNSS measurements are in the \( n \) coordinate system, the left-invariant error form can be directly represented:
\begin{equation}\label{GNSS update}
\begin{aligned}
    \boldsymbol{z}_{v,p} &= \boldsymbol{\chi} b + w_t \\
    \left[ \begin{array}{c}
        \hat{\mathbf{v}}^n \,|\, \hat{\mathbf{p}}^n \\
        \hline
        \bm{I}_2 \\

    \end{array}\right] &= 
    \left[ \begin{array}{c|c}
        \mathbf{R}^n_b & \mathbf{v}^n \,|\, \mathbf{p}^n\\
        \hline
        \bm{0}_{2\times3} & \bm{I}_2\\
    \end{array}\right] 
    \left[\begin{array}{cc}
        \bm{0}_{3\times3} & \bm{0}_{3\times3} \\
        \hline
         1& 0 \\
         0&1\\
    \end{array} \right] + w_t
    \end{aligned}
\end{equation}

The measurement innovation \( \bm{\gamma}^L_{v,p} \) can be expressed as follows:
\begin{equation}
    \bm{\gamma}^L_{v,p} = \boldsymbol{\chi}^{-1} \boldsymbol{z}^L_{v,p} - b
\end{equation}
that is,
\begin{equation}
\begin{aligned}\label{GNSS left error update}
\bm{\gamma}^L &\approx \boldsymbol{\xi}^{L \wedge} b + \hat{\boldsymbol{\chi}}^{-1} w  
= \mathbf{H}^L_{v,p} \boldsymbol{\xi}^{L} + \hat{\boldsymbol{\chi}}^{-1} w \\
\end{aligned}
\end{equation}
in (\ref{GNSS left error update}), \( \mathbf{H}^L_{v,p} \) represents the observation matrix for the left invariant error equation, which is \textit{trajectory-independent}. Since we are using a right-invariant error description, \( \mathbf{H}^L_{v,p} \) needs to be transformed into \( \mathbf{H}^R_{v,p} \) through the adjoint matrix. According to (\ref{adjoint}), \( \boldsymbol{\xi}^R = \Adjx \boldsymbol{\xi}^L \), substituting this into (\ref{GNSS left error update}) gives:
\begin{equation}
    \bm{\gamma}^L = \mathbf{H}^L_{v,p} \Adjx^{-1} \boldsymbol{\xi}^R + \hat{\boldsymbol{\chi}}^{-1} w
\end{equation}
So we have the following transformation
\begin{equation}
    \mathbf{H}^R_{v,p} = \mathbf{H}^L_{v,p} \operatorname{\bm{A}d}_{\hat{\boldsymbol{\chi}}}^{-1}
\end{equation}
where, 
\begin{equation}
    \Adjx^{-1} = \left[
    \begin{array}{cccc}
      {\mathbf{R}^n_b}^{\top}   & \boldsymbol{0}_{3,3} & \cdots & \boldsymbol{0}_{3,3}\\
       -{\mathbf{R}^n_b}^{\top} [\mathbf{v}^n]_{\times}  & {\mathbf{R}^n_b}^{\top} & \cdots &\boldsymbol{0}_{3,3}\\
       \vdots & \vdots & \ddots & \vdots\\
       -{\mathbf{R}^n_b}^{\top} [\mathbf{v}^n_w]_{\times} & \boldsymbol{0}_{3,3}& \cdots & {\mathbf{R}^n_b}^{\top} \\
    \end{array} \right]
\end{equation}
then, 
\begin{equation}{\label{GNSS_HR_VP}}
    \mathbf{H}^R_{v,p} = {\mathbf{R}^n_b}^{\top} \left[ \begin{array}{ccc|c}
        -[\mathbf{v}^n]_{\times} & \boldsymbol{I}_{3\times3} & \boldsymbol{0}_{3\times3} &\boldsymbol{0}_{3\times3}   \\
        -[\mathbf{p}^n]_{\times} & \boldsymbol{0}_{3\times3} & \boldsymbol{I}_{3\times3} &\boldsymbol{0}_{3\times3}  \\ 
    \end{array} \right ]
\end{equation}

According to  (\ref{GNSS left error update}) and (\ref{GNSS_HR_VP}), we have
\begin{equation}\label{GNSS right error update}
\begin{aligned}
    \bm{\gamma}^L &=  \mathbf{H}^{R'}_{v,p} \boldsymbol{\xi}^R + \hat{\boldsymbol{\chi}}^{-1} w\\
    \left[
    \setlength{\arraycolsep}{2pt}
    \begin{array}{c}
          \hat{v}^n - v^n  \\
          \hat{p}^n - p^n 
    \end{array} \right] 
    &= \left[ 
    \setlength{\arraycolsep}{2pt}
    \begin{array}{ccc|c}
        -[\mathbf{v}^n]_{\times} & \boldsymbol{I}_{3,3} & \boldsymbol{0}_{3,3} &\boldsymbol{0}_{3,12}  \\
        -[\mathbf{p}^n]_{\times} & \boldsymbol{0}_{3,3} & \boldsymbol{I}_{3,3} &\boldsymbol{0}_{3,12}   \\ 
    \end{array} \right ]\boldsymbol{\xi}^R + \hat{\boldsymbol{\chi}}^{-1} w\\
\end{aligned}
\end{equation}
The $\mathbf{H}^{R'}_{v,p}$ is the observation matrix for the right invariant error update in GNSS.
\subsection{Airspeed Tube}
The airspeed sensor measures the magnitude airspeed in  \(a\) frame, and its error model is established as a Gaussian white noise model.
\begin{equation}
    \hat{v}_{TAS}^a = v_{TAS}^a + \boldsymbol{n}_{TAS},\boldsymbol{n}_{TAS}^a \sim \mathcal{N} \left( 0,\sigma_{TAS}^2\right)
\end{equation}

The relationship between the true airspeed \(\mathbf{v}_{TAS}^a\) and ground speed $\mathbf{v}^n$ is shown in (\ref{vTAS measu update}).
\begin{equation}\label{vTAS measu update}
    \mathbf{v}_{TAS}^a = \mathbf{R}^{b \top}_a \mathbf{R}^{n \top}_b \left( \mathbf{v}^n - \mathbf{v}_{wind}^n \right)
\end{equation}

Therefore, $\mathbf{v}^n$ can be solved for using the attuide, $\mathbf{v}^n$ and $\mathbf{v}_{wind}^n$.
\begin{equation}\label{vn by TAS}
    \mathbf{v}^n = \mathbf{R}^b_n \mathbf{R}^a_b\mathbf{v}_{TAS}^a + \mathbf{v}_{wind}^n
\end{equation}

Based on (\ref{GNSS right error update}), similar results can be obtained. However, the $\mathbf{v}^n$ calculated from the airspeed is coupled with \(\mathbf{R}^a_b\) and \(\mathbf{v}_{wind}^n\), so the Jacobian for (\ref{vn by TAS}) needs to be calculated separately.
\begin{equation}
\begin{aligned}
    \frac{\partial \delta \mathbf{z}}{\partial \delta \boldsymbol{\theta}^a} &= \hat{\mathbf{R}}^{b \top}_n \hat{\mathbf{R}}^{b \top}_a \mathbf{v}_{TAS}^a \\
    \frac{\partial \delta \mathbf{z}}{\partial \delta \mathbf{v}^n_{w}} &= \bm{I}_3 
\end{aligned}
\end{equation}
then, airspeed update equation can be written as,
\begin{equation}\label{TAS_H_v}
\begin{aligned}
    \left[
    \setlength{\arraycolsep}{2pt}
    \begin{array}{c}
          \hat{v}^n - v^n  \\
    \end{array} \right] 
    &=  \mathbf{H}^R_{v_A}
    \boldsymbol{\xi}^R + \hat{\boldsymbol{\chi}}^{-1} w\\
    \mathbf{H}^R_{v_A} &= \left[ 
    \setlength{\arraycolsep}{2pt}
    \begin{array}{ccccc}
    -[\mathbf{v}^n]_{\times} & \boldsymbol{I}_{3\times3} & \boldsymbol{0}_{3\times9} &  \frac{\partial \delta \mathbf{z}}{\partial \delta \boldsymbol{\theta}^a} & \frac{\partial \delta \mathbf{z}}{\partial \delta \mathbf{v}^n_{w}}\\
    \end{array} \right ]
\end{aligned}
\end{equation}
The matrix $\mathbf{H}^R_{v_A}$ is the observation matrix derived from the airspeed update equation.
\subsection{Barometer}
The barometer measures the static pressure $P_s$ in the air, and its simplified conversion equation to altitude is given by (\ref{pressure2height}):
\begin{equation}\label{pressure2height}
    H_b = 44300\times \left[ 1-\left( \frac{P_s}{P_0}\right)^{\frac{1}{5.255}}\right]
\end{equation}
here, $H_b$ represents the absolute altitude, $P_0 = 1.01325 \text{bar}$ is the standard atmospheric pressure. The measurement error of the barometer is mainly influenced by temperature and airspeed. Since the barometer measures pressure altitude, to unify the coordinate system, pressure altitude is converted to the $n$ frame. Due to the significant impact of temperature fluctuations, temperature calibration should be applied in advance to improve altitude measurement accuracy while maintaining generality. The error model of the barometer, after temperature compensation, is established as Gaussian white noise, as shown in (\ref{baro update}).
\begin{equation}\label{baro update}
    \hat{\boldsymbol{H}}^n_b = \boldsymbol{H}^n_b + \boldsymbol{n}_b,\boldsymbol{n}_b \sim \mathcal{N} \left( 0,\sigma_{H_b}^2\right)
\end{equation}

Since the barometer is consistent with GNSS in the coordinate system and is one-dimensional, (\ref{GNSS left error update}) can be rewritten as follows:
\begin{equation}\label{Baro_H_h}
    \begin{aligned}
    \bm{\gamma}^L &=  \mathbf{H}^{R'}_{H_b} \boldsymbol{\xi}^R + \hat{\boldsymbol{\chi}}^{-1} w\\
          \hat{H}^n_b - H^n_b
    &= \boldsymbol{A}_b\left[ 
    \setlength{\arraycolsep}{2pt}
    \begin{array}{ccc|c}
        -[\mathbf{p}^n]_{\times} & \boldsymbol{0}_{3\times3} & \boldsymbol{I}_{3\times3} &\boldsymbol{0}_{3\times12}   \\ 
    \end{array} \right ]\boldsymbol{\xi}^R \\
    &+ \boldsymbol{A}_b\hat{\boldsymbol{\chi}}^{-1} w\\
\end{aligned}
\end{equation}
where $\boldsymbol{A}_b = 
\setlength{\arraycolsep}{2pt}
\left[\begin{array}{ccc}
    0 & 0 & 1  \\
\end{array} \right]$ is the dimension reduction matrix.

\subsection{Magnetometer}
The magnetometer measures the magnitude of the three-axis magnetic field in the ${b}$ frame, and it needs to undergo hard iron calibration before use. It is also modeled as a white noise model:
\begin{equation}\label{Magn update}
    \hat{\boldsymbol{m}} = \hat{\mathbf{R}}_n^b \mathbf{e}_1
    + \boldsymbol{n}_m,\boldsymbol{n}_m \sim \mathcal{N} \left( 0,\sigma_{m_b}^2\right)
\end{equation}
where, $\mathbf{e}_1 = \left[
\setlength{\arraycolsep}{2pt}
\begin{array}{ccc}
1 & 0 & 0 \
\end{array}\right]^{\top}$.
\begin{equation}
    \mathbf{e}_{1}^n = {\mathbf{R}^b_n}\hat{\mathbf{m}} - \boldsymbol{n}_m
\end{equation}
It can be easily built on left invariant updates:
\begin{equation}
    \left[\begin{array}{c}
         \mathbf{e}^n_1  \\
         \hline
         \bm{0}_{2\times1} \\
    \end{array}  \right]  =  \left[ \begin{array}{c|c}
        \mathbf{R}^n_b & \mathbf{v}^n \,\,\, \mathbf{p}^n\\
        \hline
        \bm{0}_{2\times3} & \bm{I}_{2}\\
    \end{array}\right] 
    \left[ \begin{array}{c}
         \hat{\mathbf{m}}  \\
         \hline
         \bm{0}_{2\times1} \\
    \end{array} \right] -w
\end{equation}

The measurement innovation $\bm{\gamma}^L_{m}$ can be expressed as follows:
\begin{equation}
    \bm{\gamma}^L_{m} = \boldsymbol{\chi}^{-1} \boldsymbol{z}^L_{m} - b
\end{equation}
that is,
\begin{equation}
    \begin{aligned}
        \bm{\gamma}^L &\approx \boldsymbol{\xi}^{L \wedge} b + \hat{\boldsymbol{\chi}}^{-1} w  
= \mathbf{H}^L_{m} \boldsymbol{\xi}^{L} + \hat{\boldsymbol{\chi}}^{-1} w \\
        \begin{array}{c}
           \mathbf{R}^b_n \mathbf{e}_1  \\
        \end{array}  &= \left[ \begin{array}{c|c}
            -[\hat{\mathbf{m}}]_{\times} &\boldsymbol{0}_{3\times18}  \\
        \end{array} \right] \delta \mathbf{x}^L +  \hat{\mathbf{R}}^b_n \mathbf{n}_m
    \end{aligned}
\end{equation}

Perform the following transformation to convert $\mathbf{H}^L_{m}$ to $\mathbf{H}^R_{m}$:
\begin{equation}
    \mathbf{H}^R_{m} = \mathbf{H}^L_{m} \operatorname{\bm{A}d}_{\hat{\boldsymbol{\chi}}}^{-1}
\end{equation}
then,
\begin{equation}\label{magn_H_m}
    \mathbf{H}^R_{m} = \left[ \begin{array}{c|c}
            [\hat{\mathbf{m}}]_{\times} &\boldsymbol{0}_{3\times18}  \\
        \end{array} \right]
\end{equation}

\subsection{Rudder Constrain}
Rudder angle (RA) is a major factor in changing the forces acting on an aircraft, while forces and moments are the direct causes affecting the aircraft's flight state. Therefore, RA is closely related to flight states such as AOA and SA. By making full use of easily accessible control inputs such as RA command, mapping between RA and AOA/SA can be established, thus applying constraints on the flow angles and improving the consistency of observations. However, accurate aerodynamic parameters of the aircraft typically require wind tunnel experiments, which are not feasible for low-cost unmanned aircraft. A common approach is to simplify the model further and identify key parameters through experiments. Nevertheless, it has been observed that due to the strong nonlinearity of the model, linear parameter identification methods may suffer from significant output noise. Therefore, this paper proposes the use of a LSTM network, a type of Recurrent Neural Network (RNN), to replace linear parameter identification in order to enhance AOA prediction accuracy under strong nonlinear conditions.

The lift force experienced by a UAV during flight is primarily influenced by lift coefficients. To estimate \(\alpha\) and \(\beta\), it is essential to determine the four major lift coefficients for the lift equation. \eqref{lift coefficient} and \eqref{yawing coefficient} provide approximate expressions for calculating these lift coefficients \cite{chao2017flight}. In these equations, \(C_{L_0}\) is the zero AOA lift coefficient, while \(C_{L_\alpha}\), \(C_{L_q}\), and \(C_{L_{\delta_e}}\) are the lift coefficients induced by the \(\alpha\), pitch rate $q$, and elevator deflection \(\delta_e\), respectively. In the conventional flight configuration of UAVs, the longitudinal and vertical planes are often decoupled for computational analysis. Due to the locally linear characteristics that UAVs exhibit for various aerodynamic coefficients within a certain control range, the equations for the vertical and lateral aerodynamic force coefficients can be simplified as follows:
\begin{equation}\label{lift coefficient}
    C_L = C_{L_0} + C_{L_\alpha} \alpha + C_{L_q} q + C_{L_{\delta_e}} \delta_e
\end{equation}
\begin{equation}\label{yawing coefficient}
C_Y =C_{Y_0} + C_{Y_\beta}\beta + C_{Y_{\delta_r}} \delta_r + C_{Y_p} p +  C_{Y_r} r
\end{equation}
The total lift force \(L\) experienced by a UAV during flight can be observed using data from an accelerometer. Based on this, an equation can be established according to the lift coefficients as follows:
\begin{equation}
    \hat{L} \approx-m a_{z}^a=\bar{q} S C_L
\end{equation}   
\begin{equation}
    \hat{Y} \approx-m a_{y}^a=\bar{q} S C_Y
\end{equation}
where, 
\begin{equation}
    \left[ 
    \begin{array}{ccc}
        a^a_{x} & a^a_{y} & a^a_{z}\\
    \end{array}
    \right]^{\top} = \mathbf{C}^a_b \hat{\mathbf{a}}^b
\end{equation}
This allows for the determination of the equivalent lift and side-force coefficients, \(\overline{C_L}\) and \(\overline{C_Y}\), respectively. Here, \(\overline{q}\) represents the equivalent dynamic pressure, \(S\) is the equivalent wing area, \(\rho\) is the air density, and \(V_{TAS}\) represents the magnitude of $\mathbf{v}_{TAS}^a$ .
\begin{equation}
    \overline{C_L}=\frac{-m a_{z}^a}{\bar{q} S},
    \overline{C_Y}=\frac{m a_{y}^a}{\bar{q} S},
    \overline{q} = \frac{1}{2} 
    \rho V_{TAS}^2
\end{equation}
\begin{figure}[tbp]
\centerline{
    \includegraphics[width=1\linewidth]{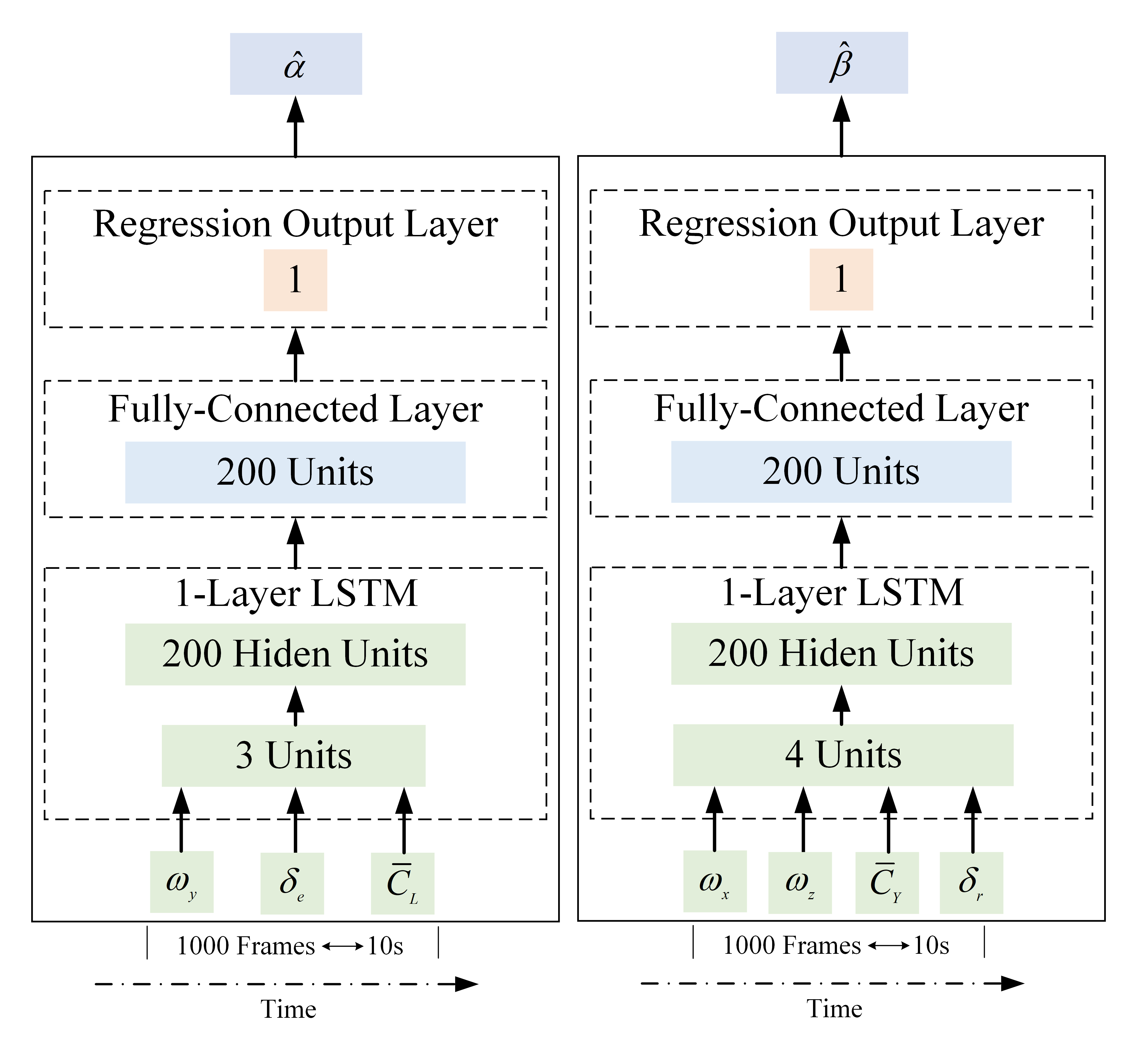}}
    \centering
    \caption{The struction of the perception model based on LSTM.}
    \label{LSTM struction}
\end{figure}
Due to the strong correlation between \((\overline{C}_L, q, \delta_e) \sim \alpha\) and \((\overline{C}_Y, \delta r, p, r) \sim \beta\), which is not only evident in the simplified models but also in sensor errors—including those from the IMU, airspeed sensors, and discrepancies between RA commands and actual feedback—using traditional least squares estimation may not yield satisfactory results (as seen in the experimental section). We thus opt for deep neural network methods for prediction.

Given that the flow angles are not solely related to sensor data at a single moment, but rather closely tied to a sequence of flight states over a period of time, we consider leveraging the generalization capability of LSTM to perform realtime estimation of \(\alpha\) and \(\beta\). The structure of this LSTM network is illustrated in Fig. \ref{LSTM struction}.
The method of estimating AOA/SA using the LSTM network is independent of velocity and position signals provided by GPS, thus allowing for tolerance against drift caused by the absence of external location signals in denied environments. The estimation relies solely on easily accessible RA command signals, angular rates and acceleration measured by IMU, as well as fundamental UAV geometrical parameters of UAV, including wingspan, wing area, weight, and so on, as illustrated in Fig. \ref{fig_UAV}. This provides an estimation reference for the UAV's flow angles even in environments where external GNSS signals are denied.

Furthermore, the network's predicted outputs for \(\alpha\) and \(\beta\) are substituted into (\ref{Rab by alpha}), integrating them into the framework of the InEKF update process. This can be seen as an advanced strategy that replaces traditional schemes based on accurate velocity, particularly in environments where GNSS is denied.
\section{Results}\label{section5}
\begin{figure*}[tbp]
\centerline{\includegraphics[width=0.8\textwidth]{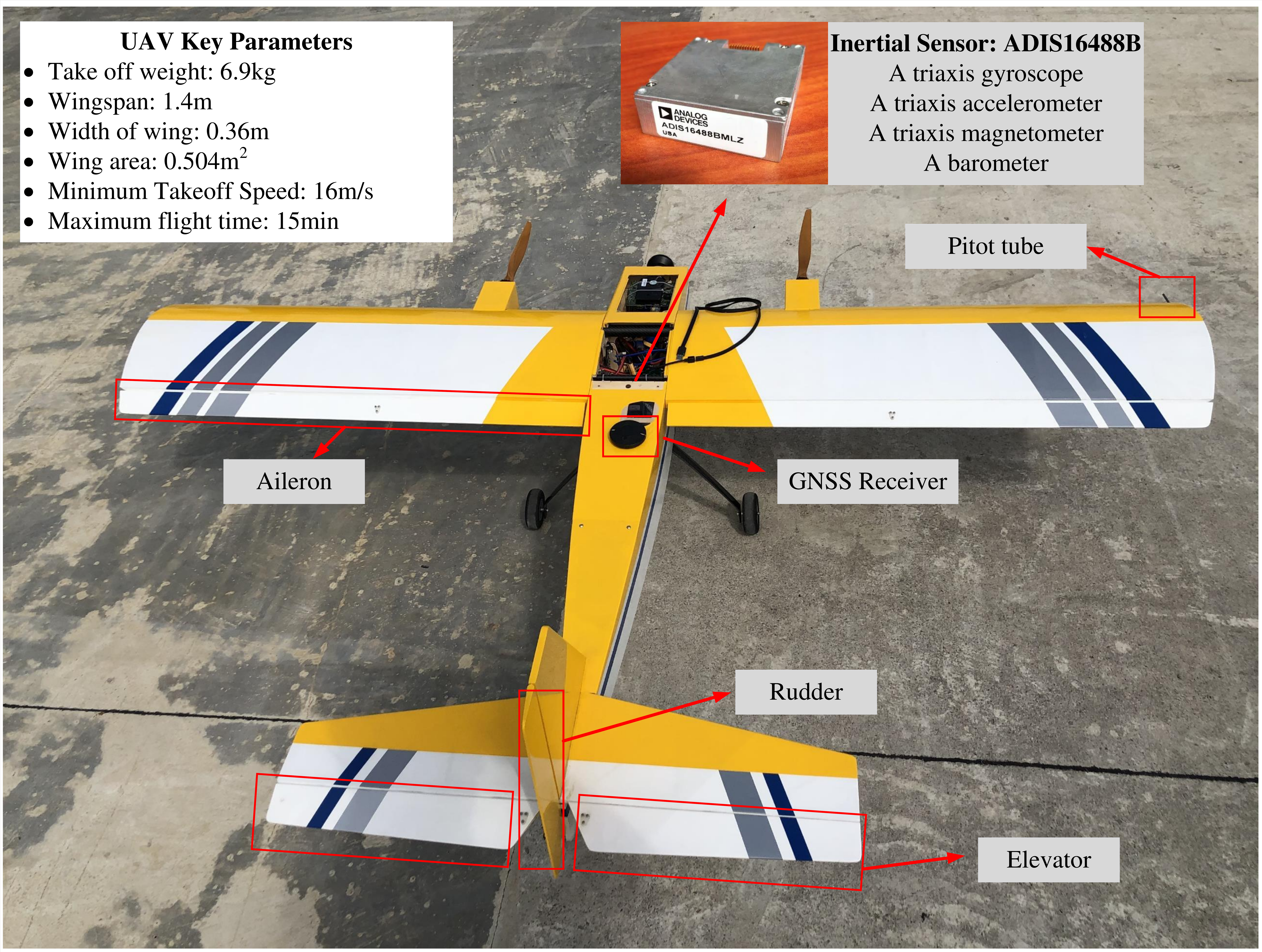}}
\caption{The fixed-wing UAV experimental platform used in our experiments is equipped with a custom industrial-grade flight controller, which includes a variety of sensor resources such as IMU, barometer, magnetometer, pitot tube, and GNSS. This UAV has a standard layout and is primarily used for GNC (Guidance, Navigation, and Control) algorithm verification. Its key ground-measurable parameters are shown in the top left corner. As a fully autonomous UAV, its takeoff and landing are algorithm-controlled, and it features a data recording system for easy data export and analysis.}
\label{fig_UAV}
\end{figure*}
To validate the effectiveness of the right-invariant multisensor fusion algorithm proposed in this paper, offline verification analysis was conducted using real flight experimental data from fixed-wing UAVs. The UAV's flight platform is illustrated in Fig.~\ref{fig_UAV}. The platform is equipped with a rich array of sensors: The flight control inertial unit uses ADIS16488 with a data sampling rate of 200Hz, which includes a three-axis gyroscope, a three-axis accelerometer, a three-axis magnetometer, and a barometer. The GNSS system employs NovAtel hardware with a sampling frequency of 5Hz. The UAV has a conventional configuration, featuring three types of control surfaces: elevators on the horizontal tail, a rudder on the vertical tail, and ailerons on the wingtips. The performance specifications and sampling frequencies of the onboard sensors for this UAV are presented in Table.~\ref{Sensor specification}. The main operational flow of the right-invariant algorithm proposed in this paper is shown in Algorithm.~\ref{ES-RIEKF process}.

The ES-RIEKF algorithm proposed in this paper is mainly compared with NCF \cite{mahony2008nonlinear,ye2022airspeed}, ES-EKF \cite{geneva2020openvins}, and ES-LIEKF. Note: For the ES-EKF, we utilize only the algorithm pipeline and do not incorporate camera data fusion. The NCF algorithm is suitable for deployment in embedded systems due to its low computational cost and efficiency, making it a preferred choice for low-cost UAV attitude estimation. ES-EKF represents a significant body of work in filter-based algorithms in recent years, while ES-LIEKF offers another geometric expression for InEKF. 

As the framework of ES-RIEKF is relatively similar to that of ES-EKF and ES-LIEKF, the same filter parameters such as \(\mathbf{P}_0, \mathbf{Q}_0, \mathbf{R}_0\) were used in both sets of programs for effective comparison. The values of these filter parameters are presented in Table.~\ref{Filter parameters}. 

The advantages of the algorithm will be evaluated from the following four core aspects.
 \begin{enumerate}[(1)]
\item ES-RIEKF exhibits superior convergence characteristics when faced with attitude disturbances.
\item A comparison of pose accuracy of ES-RIEKF with mainstream algorithms such as ES-EKF and NCF.
\item Utilization of LSTM networks to fuse readily available sensor data like RA for predicting AOA/SA flow angles.
\item As a multisensor navigation algorithm, the resilience advantages it offers when facing partial sensor failures.
\end{enumerate}

\begin{table*}[tbp]\caption{Specifications of Multisensors on the fix-wing UAV}
    \centering
    \begin{tabular}{ccccc}
    \hlineB{2}
       \textbf{Source}  & 
       \textbf{Measurements}  & 
       \textbf{Frequency} & 
       \textbf{Parameters} & 
       \textbf{Typical Value}\\
    \hline
    \multirow{4}{*}{IMU} & 
    \multirow{2}{*}{$\bm{\omega}^b$} &
    \multirow{4}{*}{$100 \text{Hz}$}& 
    In-Run Bias Stability of Gyroscope & 
    $6.25^{\circ}/\text{hr}$\\
        & & &
        Angular Random Walk & $0.3^{\circ}/\sqrt{\mathrm{hr}}$  \\
        & \multirow{2}{*}{$\bm{a}^b$} & & 
        In Run Bias Stability of Accelerometer & $0.1\mathrm{m}/\mathrm{s}^2$\\
        & & & Velocity Random Walk & $0.029\mathrm{m}/\mathbf{s}/\sqrt{\mathrm{hr}}$\\ 
        \hline
    \multirow{2}{*}{GNSS} & 
        $\mathbf{p}^n_G$ &
    \multirow{2}{*}{$5$Hz} &
   Position Noise & $0.1$m \\
        &
        $\mathbf{v}^n_G$ &
        & 
       Velocity Noise &
        $0.1\mathrm{m}/\mathrm{s}$\\
    \hline
    Magnetometer & $\bm{m}$& 100Hz & Output Noise& $0.45\mathrm{mgauss}$\\
    \hline
    Barometer & $P_s$& 100Hz & Output Noise & $0.025\mathrm{mbar}$\\
    \hline
    Pitot Tube &  $V_{TAS} $ &100Hz & Output Noise & $1\mathrm{m}/\mathrm{s}$\\
    \hline
    Rudder Command & $\bm{\delta}_{aer}$& 100Hz & Output Noise & $0.1^{\circ}$\\
    \hlineB{2}
    \end{tabular}
    \label{Sensor specification}
\end{table*}

\begin{table}[tbp]
    \caption{Filter parameters setting.}
    \centering
    \begin{tabular}{clcc}
    \hlineB{2}
       \textbf{Variables}& \textbf{Parameters}&\textbf{Values}  & \textbf{Units}\\
       \hline
       \multirow{7}{*}{$\mathbf{Q}_0$} 
       & $\sigma_a^2$ & $(10^{-3})^2$ & $\left(\mathrm{m}/\mathrm{s}^2\right)^2$\\
       & $\sigma_g^2$ & $(10^{-4})^2$ & $\left(\mathrm{rad}/\mathrm{s}\right)^2$\\
       & $\sigma_{\mathbf{b}_a}^2$ & $(10^{-4})^2$ & $\left(\mathrm{m}/\mathrm{s}^2\right)^2/\mathrm{Hz}$ \\
       & $\sigma_{\mathbf{b}_g}^2$ & $(2\times10^{-5})^2$ & $\left(\mathrm{rad}/\mathrm{s}\right)^2/\mathrm{Hz}$ \\
       & $\sigma_{\mathbf{b}_m}^2$ & $(2\times10^{-2})^2$ & $\left(\mathrm{mguass}\right)^2/\mathrm{Hz}$ \\
       & $\sigma_{\delta \boldsymbol{\theta}_a}^2$ & $(0.0175)^2$ & $\left(\mathrm{rad}\right)^2$ \\
       & $\sigma_{\delta \mathbf{v}_{wind}}^2$ & $(0.01)^2$ & $\left(\mathrm{m}/\mathrm{s}\right)^2$ \\
       \hline
       \multirow{8}{*}{$\mathbf{P}_0$}
       & $\sigma_{\delta \boldsymbol{\theta}_I}^2$ & $([1;1;5]/57.3)^2$ & $\mathrm{rad}^2$\\
       & $\sigma_{\delta \mathbf{v}}^2$ & $(0.1)^2$ & $(\mathrm{m}/\mathrm{s})^2$\\
       & $\sigma_{\delta \mathbf{p}}^2$ & $(0.2)^2$ & $\mathrm{m}^2$\\
       & $\sigma_{\delta \mathbf{b}_g}^2$ & $(4.8478e-5)^2$ & $(\mathrm{rad}/\mathrm{s})^2$\\
       & $\sigma_{\delta \mathbf{b}_a}^2$ & $(0.05)^2$ & $(\mathrm{m}/\mathrm{s}^2)^2$\\
       & $\sigma_{\mathbf{b}_m}^2$ & $(2\times10^{-2})^2$ & $\left(\mathrm{mguass}\right)^2/\mathrm{Hz}$ \\
       & $\sigma_{\delta \boldsymbol{\theta}_a}^2$ & $(0.035)^2$ & $\left(\mathrm{rad}\right)^2$ \\
       & $\sigma_{\delta \mathbf{v}_{wind}}^2$ & $(0.1)^2$ & $\left(\mathrm{m}/\mathrm{s}\right)^2$ \\
       \hline
       \multirow{5}{*}{$\mathbf{R}_0$}
       & $\sigma_{\mathbf{v}_G}^2$ & $(0.01)^2$  & $(\mathrm{m}/\mathrm{s})^2$\\
       & $\sigma_{\mathbf{p}_G}^2$ & $(0.1)^2$ & $(\mathrm{m})^2$\\
       & $\sigma_{\mathbf{v}_{TAS}}^2$ & $(0.1)^2$ & $(\mathrm{m}/\mathrm{s})^2$\\
       & $\sigma_{\mathbf{H}_{baro}}^2$ & $(0.1)^2$ & $(\mathrm{m})^2$\\
       & $\sigma_{\mathbf{m}}^2$ & $(0.1)^2$ & $(\mathrm{mguass})^2$\\
       \hlineB{2}
    \end{tabular}
    \label{Filter parameters}
\end{table}

\begin{algorithm*}
\caption{ES-RIEKF Algorithmic Processing}
\label{ES-RIEKF process}
\textbf{Initialization:} Initialize filter parameters: \( \bm{\chi}_0, \mathbf{P}_0, \mathbf{Q}_0, \mathbf{R}_0, \mathbf{G}_0, \mathbf{g}_0 \) \\
\While{Receiving multi-source sensor data}{
    \If{Valid \textbf{IMU} data is received}{
        \textbf{Time Update:}\\
        \hspace{\algorithmicindent} SINS Update: \( \mathbf{R}_{t+1}, \mathbf{v}_{t+1}, \mathbf{p}_{t+1} \leftarrow \mathbf{R}_{t}, \mathbf{v}_{t}, \mathbf{p}_{t} \) 
        (refer to (\ref{state update dynamic}))\\
        \hspace{\algorithmicindent} Compute Error Transition Matrix: \( \boldsymbol{\Phi}_{t,t+1} \leftarrow \exp(\mathbf{F}_t) \) \\
        \hspace{\algorithmicindent} Right-Invariant Error State One-Step Prediction: \( d\boldsymbol{\chi}^R_{t,t+1} \leftarrow \boldsymbol{\Phi}_{t,t+1} d\boldsymbol{\chi}^R_{t} \) \\
        \hspace{\algorithmicindent} One-Step Prediction Covariance Update: \( \mathbf{P}_{t,t+1} \leftarrow \boldsymbol{\Phi}_{t,t+1} \mathbf{P}_{t} \boldsymbol{\Phi}_{t,t+1}^{\top} + \mathbf{G}_0 \mathbf{Q}_0 \mathbf{G}_0^{\top} \)
    }
    \If{Valid \textbf{GNSS} signal \( \mathbf{v}_G^n, \mathbf{p}_G^n \) is received}{
        \textbf{GNSS Measurement Update:}\\
        \hspace{\algorithmicindent} Compute Right-Invariant Observation Matrix: \( \mathbf{H}^{R}_{t+1} \leftarrow \mathbf{H}^R_{v,p} \) (refer to  (\ref{GNSS_HR_VP}))\\
        \hspace{\algorithmicindent} Compute Observation Vector: \( \mathbf{z}_k \leftarrow [\mathbf{v}_I - \mathbf{v}_G; \mathbf{p}_I - \mathbf{p}_G] \) \\
        \hspace{\algorithmicindent} Compute KF Gain: \( \mathbf{K}_{t+1} \leftarrow \mathbf{P}_{t,t+1} \mathbf{H}^{R \top}_{t+1} (\mathbf{H}^{R}_{t+1} \mathbf{P}_{t,t+1} \mathbf{H}^{R\top}_{t+1})^{-1} \) \\
        \label{step1}
        \hspace{\algorithmicindent} State Update: \( d\boldsymbol{\chi}^R_{t+1} \leftarrow d\boldsymbol{\chi}^R_{t,t+1} + \mathbf{K}_{t+1} (\mathbf{z}_k - \mathbf{H}^{R}_{t+1} d\boldsymbol{\chi}^R_{t,t+1}) \) \\
        \hspace{\algorithmicindent} Covariance Update: \( \mathbf{P}_{t+1} \leftarrow (\mathbf{I} - \mathbf{K}_{t+1} \mathbf{H}^{R}_{t+1}) \mathbf{P}_{t,t+1} \)
        \label{step2}
    }
    \If{Valid \textbf{barometer} \( P_s \) is received}{
    \textbf{Barometer Measurement Update:}\\
    \hspace{\algorithmicindent} Compute Right-Invariant Observation Matrix: \( \mathbf{H}^{R}_{t+1} \leftarrow \mathbf{H}^{R'}_{H_b} \) (refer to  (\ref{Baro_H_h})) \\
    \hspace{\algorithmicindent} Compute Observation Vector: \( \mathbf{z}_k \leftarrow [H_I - H_b] \) \\
    Repeat from \textbf{Step}~\ref{step1} to \textbf{Step}~\ref{step2}
}
    \If{Valid \textbf{magnetometer} \( \bm{m} \) is received}{
    \textbf{Magnetometer Measurement Update:}\\
    \hspace{\algorithmicindent} Compute Right-Invariant Observation Matrix: \( \mathbf{H}^{R}_{t+1} \leftarrow \mathbf{H}^R_{m} \) (refer to  (\ref{magn_H_m})) \\
    \hspace{\algorithmicindent} Compute Observation Vector: \( \mathbf{z}_k \leftarrow \left[\mathbf{m}_b \right] \) \\
    Repeat from \textbf{Step}~\ref{step1} to \textbf{Step}~\ref{step2}
}
    \If{Valid \textbf{airspeed} \( \mathbf{v}_{\text{TAS}}^a \) is received}{
    \textbf{True Airspeed Measurement Update:}\\
    \hspace{\algorithmicindent} Compute Right-Invariant Observation Matrix: \( \mathbf{H}^{R}_{t+1} \leftarrow \mathbf{H}^R_{v_A} \) (refer to  (\ref{TAS_H_v})) \\
    \hspace{\algorithmicindent} Compute Observation Vector: \( \mathbf{z}_k \leftarrow \left[\mathbf{v}_I - \mathbf{v}_G \right] \) \\
    Repeat from \textbf{Step}~\ref{step1} to \textbf{Step}~\ref{step2}
}
    Output predicted error state \( \delta \mathbf{x}_{t+1,t+1} \) and covariance \( \mathbf{P}_{t+1,t+1} \) \\
    Reset error state: \( \delta \mathbf{x}_{t+1,t+1} \leftarrow \bm{0} \) \\
    Error feedback: $\hat{\boldsymbol{\omega}} \longleftarrow \boldsymbol{\omega}_{raw} - \sum_{t=0}^t \delta \mathbf{b}_g, \hat{\boldsymbol{a}} \longleftarrow \boldsymbol{a}_{raw} - \sum_{t=0}^t \delta \mathbf{b}_a$,$\bm{\chi} \longleftarrow \hat{\bm{\chi}} \boxplus \delta \mathbf{x}$ (refer to  (\ref{RInEKF def}))
    }
\end{algorithm*}

\subsection{Convergence}
State consistency is primarily reflected in the aircraft's convergence ability and the consistency of state estimation under unknown disturbances. Previous works has mostly been based on simulation environments for comparative analysis. To enhance credibility and investigate real-world application effects, this section uses actual flight data from UAVs to compare the convergence characteristics of ES-EKF and ES-RIEKF under initial attitude disturbances. Initially, we introduce disturbances into the core attitude of the navigation system at the moment of takeoff, using the same measurement values, initial covariance, and measurement noise covariance matrices. Specifically, we impose a uniform bias of \( -30^\circ \) to \( 30^\circ \) on the roll and pitch angles to observe the impact of different disturbances on the accuracy and convergence speed of attitude estimation.

\begin{figure*}[tbp]
\centerline{\includegraphics[width=1\textwidth]{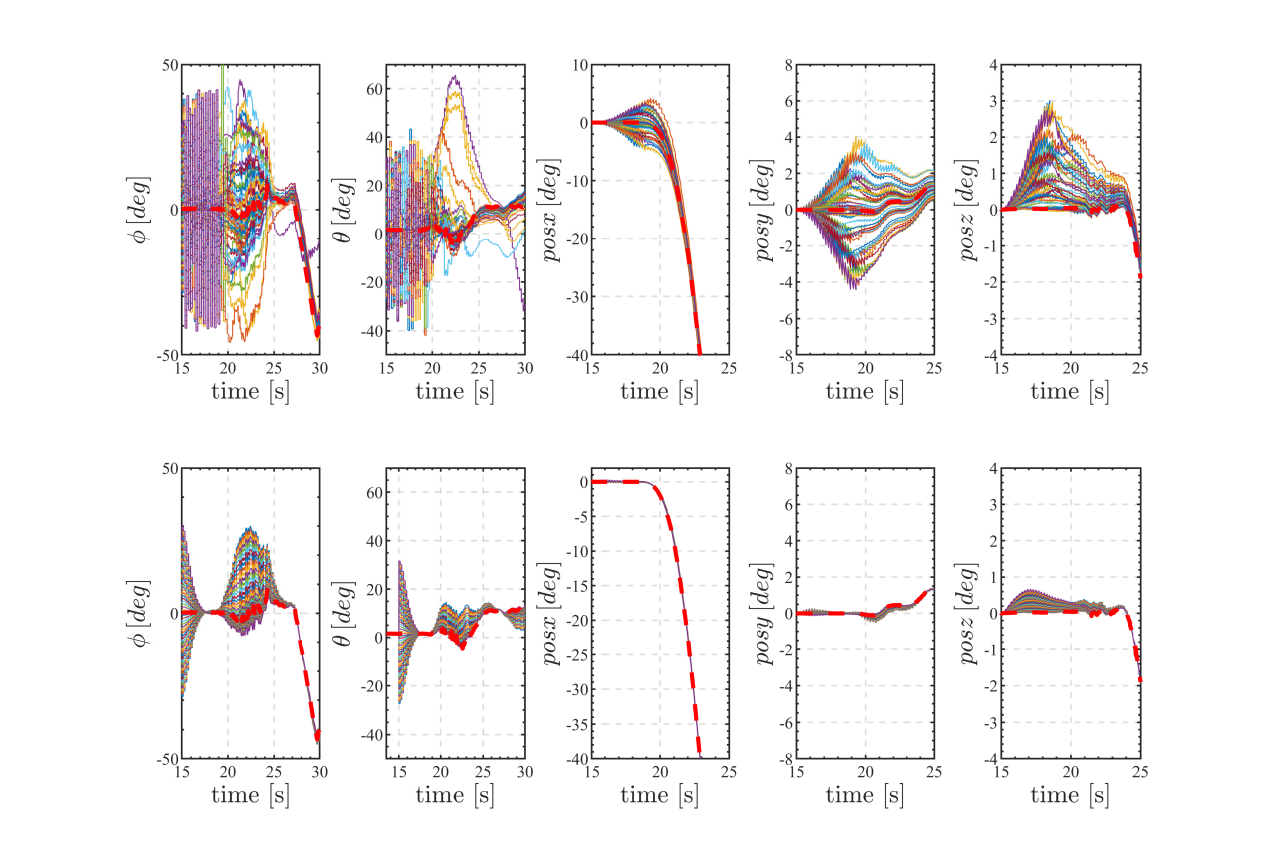}}
\caption{Comparing the convergence of ES-EKF and ES-RIEKF algorithms under attitude disturbances using the same filter parameters.}
\label{attdeviation}
\end{figure*}

The filter responses are shown in the Fig. \ref{attdeviation}. Under the same filter parameter settings, it is evident that the convergence speed of ES-RIEKF is much faster than that of ES-EKF. The ES-EKF experiences attitude oscillations when faced with large angular deviations. Around 20 seconds, the UAV starts taxiing from a stationary position, and around 25 seconds, it enters flight mode. ES-RIEKF's attitude has already converged before taxiing; however, the IMU bias estimation is influenced by the large attitude deviation, leading to a certain degree of overshooting in the attitude estimation. After the taxiing motion excitation, the attitude rapidly converges. On the other hand, the attitude error of ES-EKF under disturbances is noticeably larger, and some curves even exhibit short-term divergence. The impact on attitude fully propagates to speed and position. Correspondingly, it can be seen that the three-axis position disturbances for ES-RIEKF are minimal, with the maximum error being less than 1 meter, and the convergence speed is significantly faster than that of ES-EKF.
\subsection{Accuracy of navigation}
\begin{figure}[tbp]
\centerline{\includegraphics[width=1\linewidth]{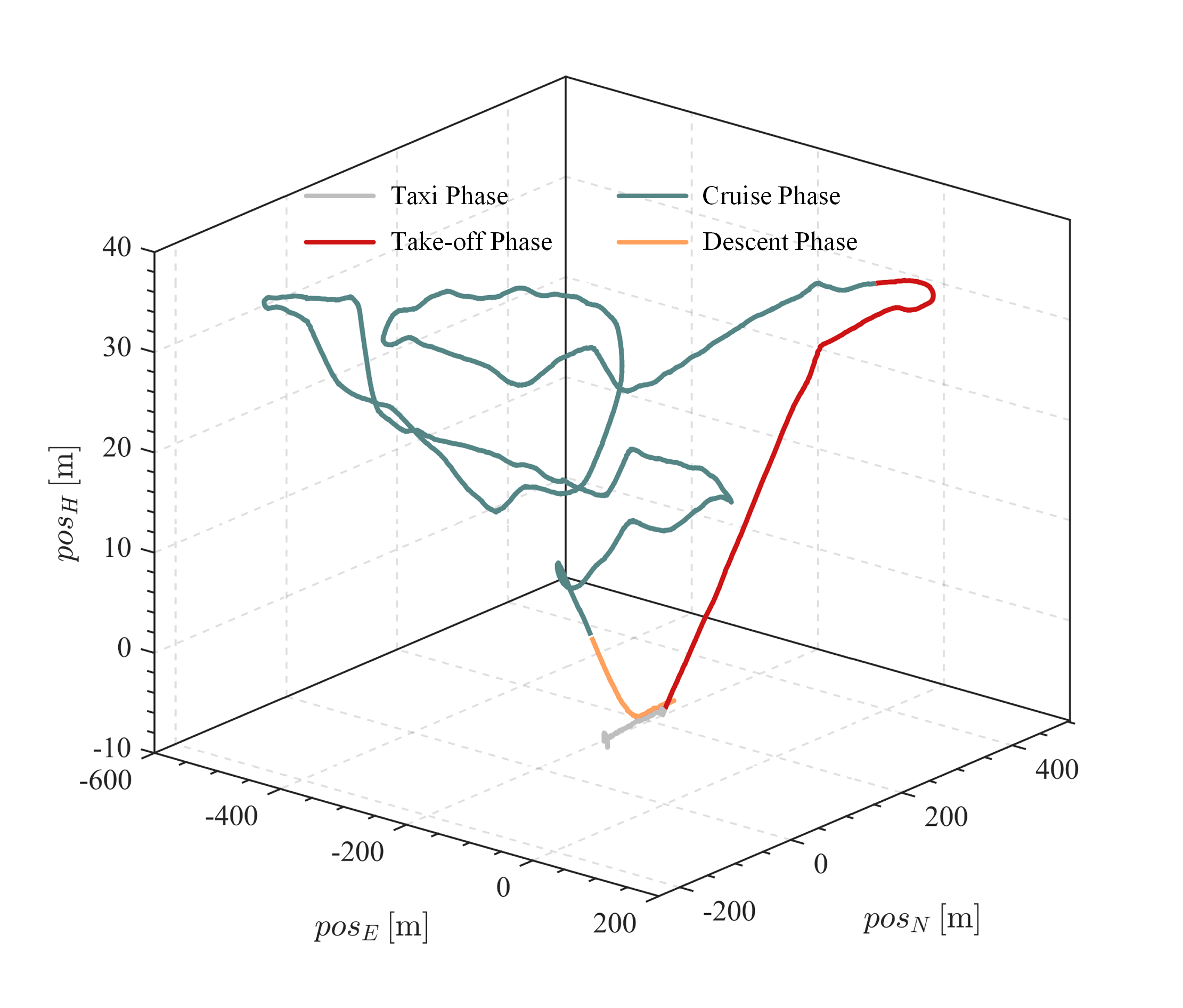}}
\caption{Flight trajectory of UAV. Different colour represent different flight phases.}
\label{3D_Traj}
\end{figure}

This section presents the estimation accuracy of different algorithms. The UAV's flight trajectory is shown in Fig. \ref{3D_Traj}. The UAV undergoes various maneuvers including taxiing, takeoff, turning, cruising, diving, and landing. Flight data is read offline and imported into relevant algorithmic simulations for analysis. We use Mean Absolute Error (MAE) and Root-Mean-Square Error (RMSE) to evaluate the algorithm performance. The definitions of MAE and RMSE are as follows:
\begin{equation}
\operatorname{MAE}(\hat{\mathbf{x}}, \mathbf{x})=\frac{1}{n} \sum_{i=1}^n\left|\hat{\mathbf{x}}_i-\mathbf{x}_i\right|
\end{equation}
\begin{equation}
\operatorname{RMSE}(\hat{\mathbf{x}}, \mathbf{x})=\sqrt{\frac{1}{n} \sum_{i=1}^n\left(\hat{\mathbf{x}}_i-\mathbf{x}_i\right)^2}
\end{equation}

\begin{figure}[tbp]
\centerline{\includegraphics[width=1\linewidth]{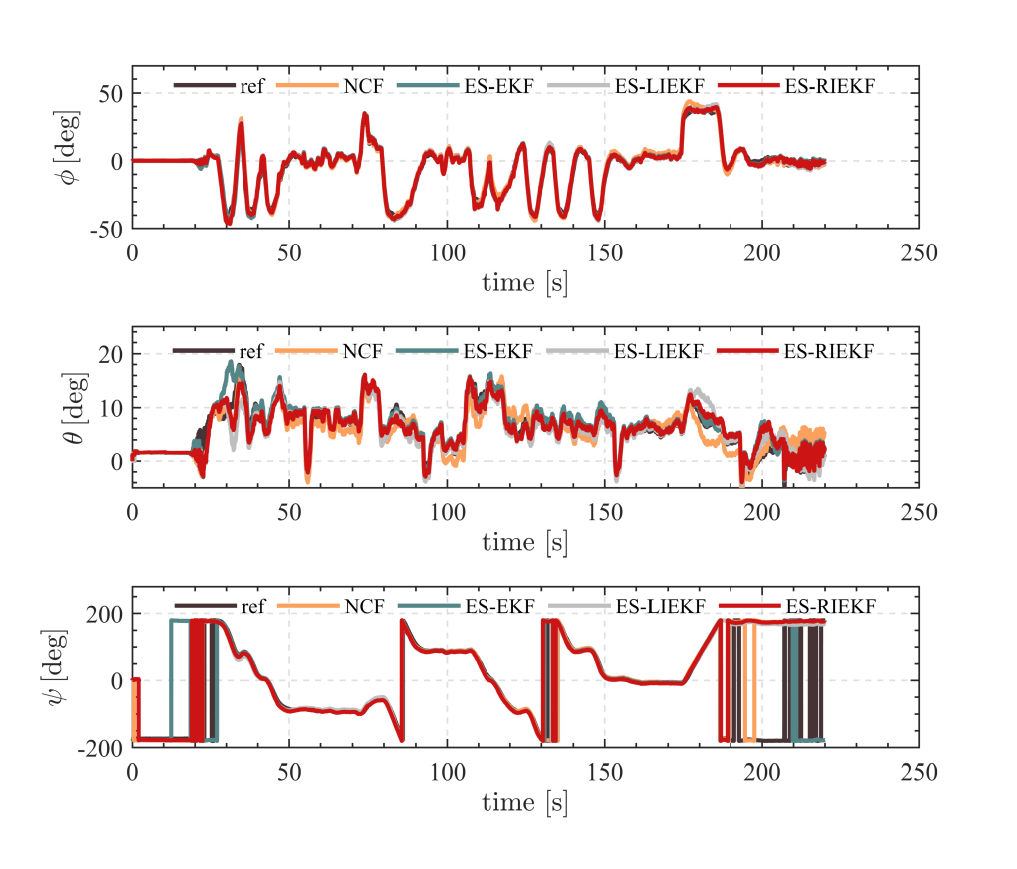}}
\caption{Comparison of attitude during flight.}
\label{att compare}
\end{figure}

Fig. \ref{att compare} shows the attitude comparison curves of different algorithms throughout the flight duration. The black curve represents the reference signal directly output by the flight control system, and the Table.~\ref{The attitude comparison} shows the statistical results of the attitude errors. From the curves, it can be seen that during the 20-30 second UAV take-off phase, ES-RIEKF converges more quickly compared to ES-EKF. In terms of overall attitude accuracy metrics, ES-RIEKF performs better in both MAE and RMSE index, showing an improvement of nearly 10\% over ES-EKF.
\begin{table}[tbp]\caption{The attitude comparison}
\centering
\begin{tabular}{lcc}
\hline 
\textbf{Algorithms} & \textbf{MAE}(deg)  & \textbf{RMSE}(deg) \\
\hline NCF & $1.7451$ & $2.2176$ \\
ES-EKF & $1.1001$ & $1.5970$ \\
ES-LIEKF & $1.2245$ & $1.8182$ \\
ES-RIEKF  & $\bm{0.9918}$ & $\bm{1.4335}$ \\
\hline
\end{tabular}
\label{The attitude comparison}
\end{table}

\begin{figure}[tbp]
\centerline{\includegraphics[width=1\linewidth]{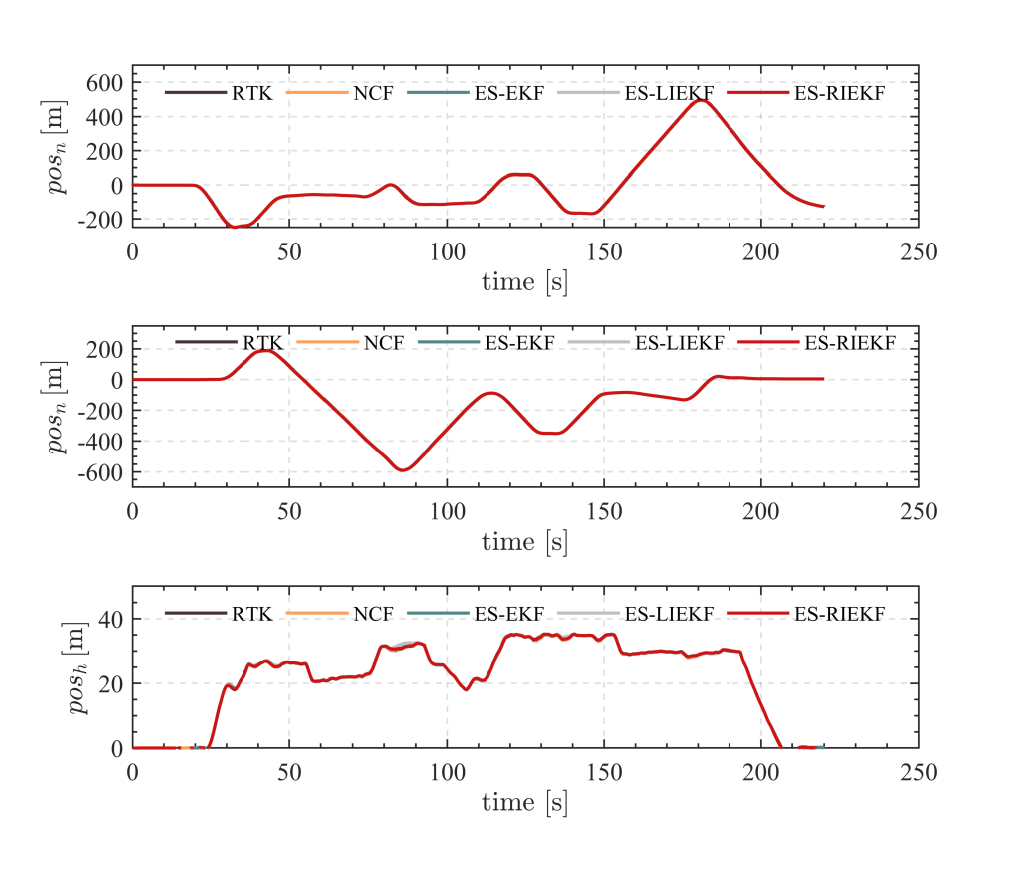}}
\caption{pos compare}
\label{pos compare}
\end{figure}

\begin{figure}[tbp]
\centerline{\includegraphics[width=1\linewidth]{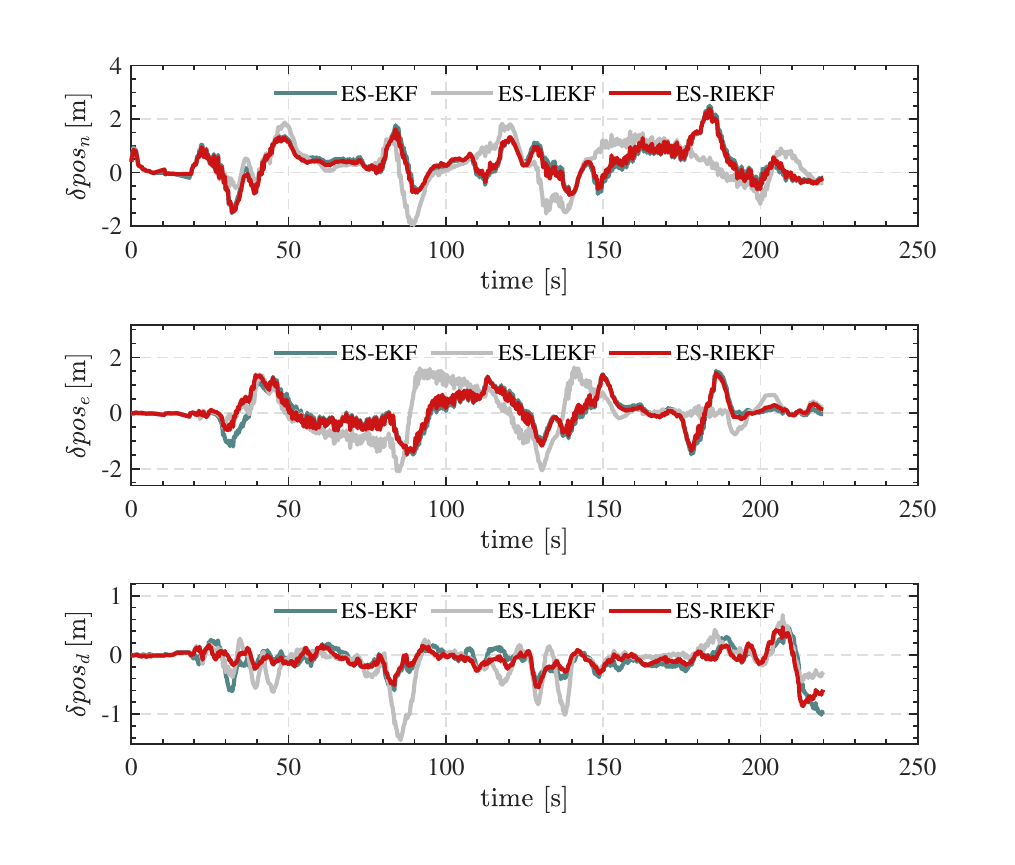}}
\caption{Comparison of position errors.}
\label{poserror compare}
\end{figure}

Fig. \ref{pos compare} and Fig. \ref{poserror compare} show the position solution curves and position error curves for different algorithms, respectively. From the position error curves, it can be seen that the overall error trends for ES-EKF and ES-RIEKF are fairly consistent. This is because their definitions of pose errors are quite similar, both being variables in the $n$ frame, as opposed to ES-LIEKF where the variable is essentially in the $b$ frame. Under the same filter parameters and sensor data preprocessing, InEKF (including both RI and EI variants) shows smaller three-axis position errors compared to ES-EKF. This is especially apparent after the excitation caused by the taxiing motion, where IEKF's errors converge more quickly. Moreover, ES-RIEKF exhibits smaller overall errors compared to ES-LIEKF, particularly in the altitude channel.
\begin{table}[tbp]\caption{The position error comparison}
\centering
\begin{tabular}{lcc}
\hline 
\textbf{Algorithms} & \textbf{MAE}(m)  & \textbf{RMSE}(n) \\
\hline 
ES-EKF & $0.385$ & $0.515$ \\
ES-LIEKF & $0.468$ & $0.624$ \\
ES-RIEKF  & $\bm{0.373}$ & $\bm{0.494}$ \\
\hline
\end{tabular}
\label{tab:my_label}
\end{table}

\begin{figure}[tbp]
\centerline{\includegraphics[width=1\linewidth]{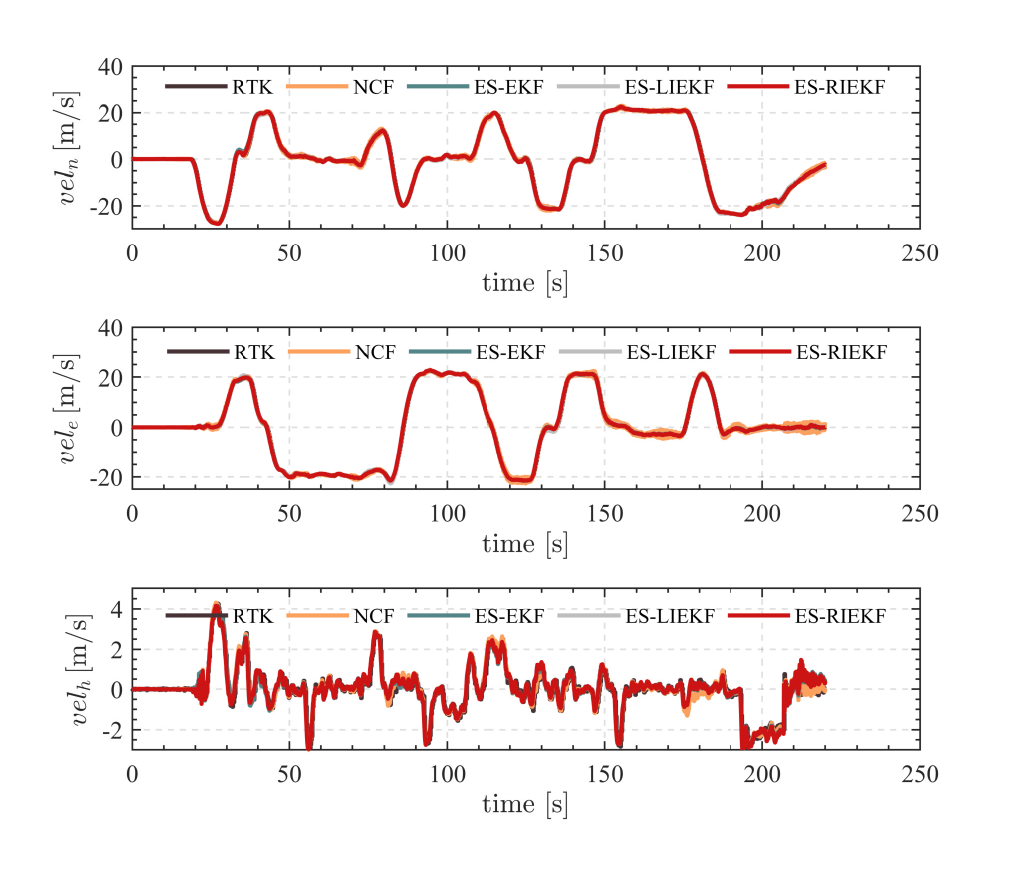}}
\caption{Comparison of velocity calculation.}
\label{vel compare}
\end{figure}

\begin{figure}[tbp]
\centerline{\includegraphics[width=1\linewidth]{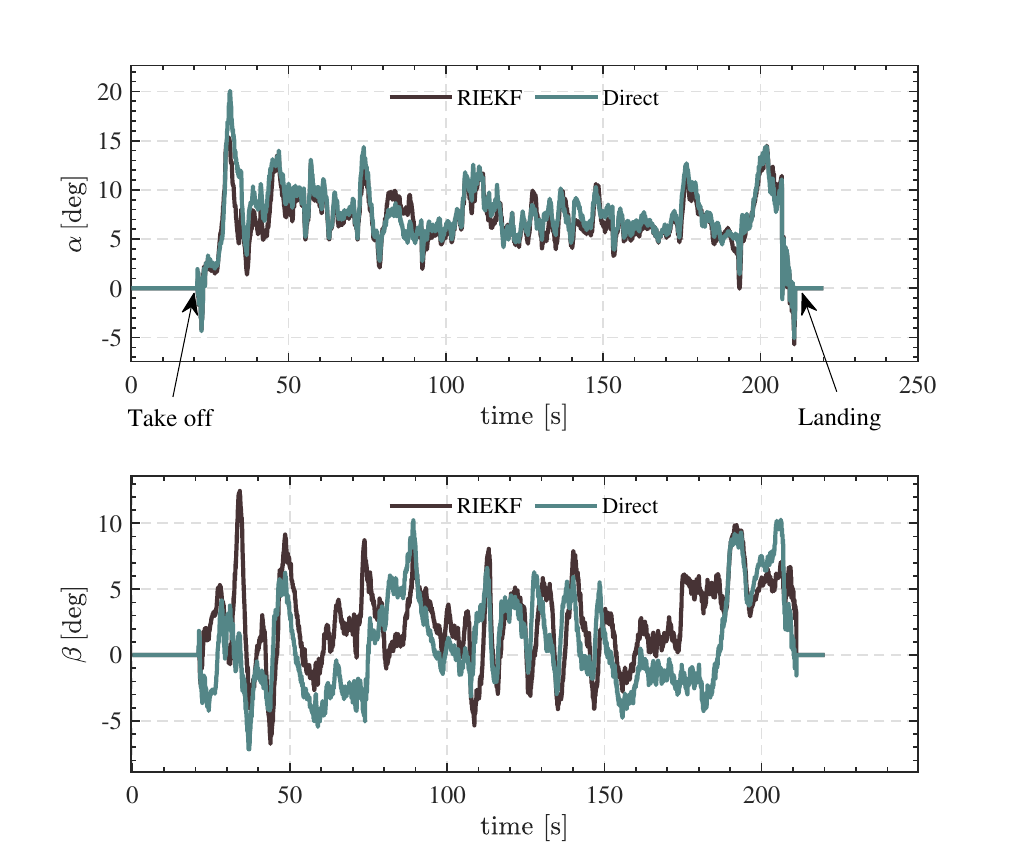}}
\caption{Comparison of AOA/SA estimation.}
\label{AOASA compare}
\end{figure}

Fig. \ref{AOASA compare} compares the estimation curves for the UAV's airflow angles AOA/SA calculated using the ES-RIEKF algorithm and a direct decoupling method \cite{durham2013aircraft}. Given that GNSS signals provide accurate speed corrections throughout the flight, the calculation of AOA and SA mainly relies on airspeed updates. Both methods generally show consistent trends; the cruising flight AOA is roughly between 5-10 degrees. During turns, due to an increase in the UAV's roll angle, some of the lift force is used as a centripetal force. Therefore, the flight control adjusts the elevator to increase the AOA and maintain altitude and speed, which is consistent with real-world observations. Compared to the fused solution, the direct calculation of AOA and SA exhibits more oscillations and deviations during turns, as it neglects the coupling between lateral and longitudinal dynamics.

\begin{figure}[tbp]
\centerline{\includegraphics[width=1\linewidth]{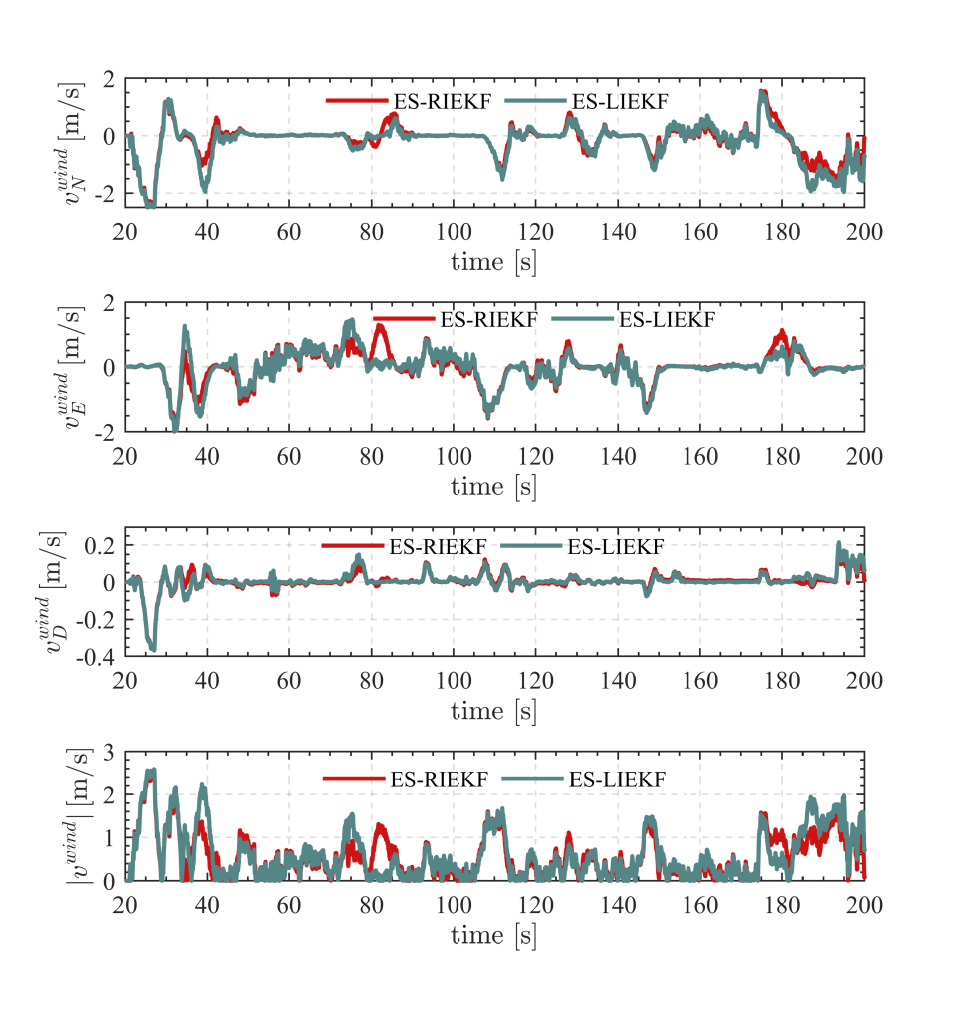}}
\caption{Comparison of wind speed estimation.}
\label{Wind compare}
\end{figure}

Fig. \ref{Wind compare} presents a comparison of wind speed predictions between the two algorithms, ES-RIEKF and ES-LIEKF. Both are able to estimate the wind field at the current moment for the UAV, with their trends being fairly consistent.

\subsection{LSTM prediction model}
\begin{figure}[tbp]
\centerline{\includegraphics[width=1\linewidth]{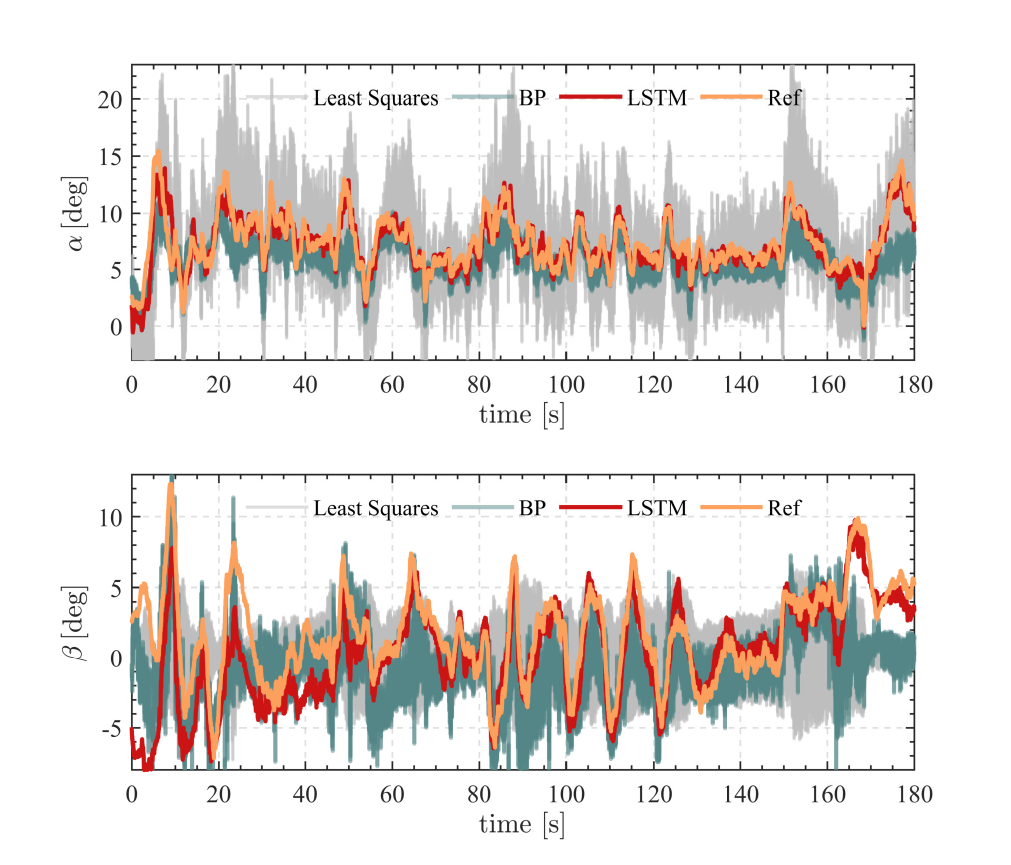}}
\caption{Comparison of Different Networks for AOA/SA Application.}
\label{AOA/SA net compare}
\end{figure}

The LSTM airflow angle prediction model is primarily designed to address the prediction issues of the AOA/SA for UAV under GNSS denied. Due to the lack of accurate speed correction from GNSS, it's challenging for UAVs to directly estimate airflow angles using kinematic models. Sophisticated dynamic models require aerodynamic parameters that low-cost UAVs can't acquire. Therefore, we followed the method in the cited literature to identify key aerodynamic parameters through least squares (LS) and then calculate the angles \( \alpha \) and \( \beta \). As shown by the gray curve in Fig. \ref{AOA/SA net compare}, the estimated AOA contains a lot of high-frequency noise. Moreover, the trend component in the SA estimation is missing at times, and the high-frequency noise would undoubtedly introduce disturbances into the system.

Subsequently, we explored other deep neural network architectures and compared a back-propagation (BP) neural network with 100 hidden layers and an LSTM neural network with 200 hidden layers. Ten sets of flight experiment data were used for training. Note that these sets were not from one UAV but multiple UAVs of the same size and type, with an overall flight time of approximately 40 minutes. We used the \( \alpha \) and \( \beta \) values obtained from ES-RIEKF algorithm, integrated with GNSS and airspeed, as the target values for training. After 500 iterations of training, it was found that the BP neural network showed improved estimation accuracy compared to least squares, reducing high-frequency oscillations while closely matching the trend in the AOA estimation. However, it failed to achieve the desired accuracy for SA prediction. The higher nonlinearity in SA compared to the AOA might make frame-by-frame prediction problematic, making it difficult for the neural network to capture finer details and instead capturing noise in the data, which negatively impacts system prediction accuracy.

In contrast, using sequence prediction, the LSTM achieves a much more desirable prediction accuracy after training with the same batch. The AOA and SA at the current moment are not only related to the current flight state but also to a short-term time sequence of flight data. Therefore, the gate structure of the LSTM allows the network to store and retrieve historical information and learn to remember patterns in the sequence, something a traditional BP neural network cannot accomplish. As can be seen, the red curve represents the LSTM's predicted values, which perfectly track the reference signal. The prediction error metrics for different networks are shown in Table \ref{AOA nn error compare}. Compared to BP and LS, the RMSE errors of \( \alpha \) and \( \beta \) predicted by LSTM are \( 0.993^\circ \) and \( 2.727^\circ \), respectively. Predicting the SA (\( \beta \)) is more challenging than the AOA (\( \alpha \)), due to its strong no-linearity and greater number of coupled factors.

\begin{table}[tbp]\caption{Comparison of Prediction Accuracy for AOA Using Different Methods.}
\setlength{\tabcolsep}{2.5pt}
\centering
\begin{tabular}{ccccc}
\hline 
& \multicolumn{2}{c}{$\boldsymbol{\alpha}$} & \multicolumn{2}{c}{$\boldsymbol{\beta}$}\\
\hline
\textbf{Algorithms} & \textbf{MAE} (deg)  & \textbf{RMSE} (deg) & \textbf{MAE} (deg)  & \textbf{RMSE} (deg)\\
\hline 
LS & $2.778$ & $3.564$ & $3.139$ & $4.004$ \\
BP & $1.701$ & $2.262$ & $2.886$ & $3.671$ \\
LSTM  & $\bm{0.711}$ & $\bm{0.993}$  & $\bm{1.776}$ & $\bm{2.727}$\\
\hline
\end{tabular}
\label{AOA nn error compare}
\end{table}

\subsection{Resilience}
Further, we evaluated the navigation and positioning accuracy of different algorithms under GNSS denied environment. To ensure consistency in the comparison, this set of experiments used the same dataset as the previous sections for off-line experiment. We chose the 90th second as the moment of GNSS denial. After that, the measurement update mode in the EKF algorithms will downgrade to using data from the barometer, magnetometer, and pitot tube sensors. The algorithms compared in this set of experiments are ES-EKF and ES-LIEKF. In the section when GNSS is denied, ES-EKF degrades to pure inertial navigation that has fully stimulated GNSS/INS maneuvers and fixed bias before the denial, serving as a control experiment. ES-RIEKF and ES-LIEKF use the same filter parameters, as shown in the Table. \ref{Filter parameters}.

\begin{figure}[tbp]
\centerline{\includegraphics[width=1\linewidth]{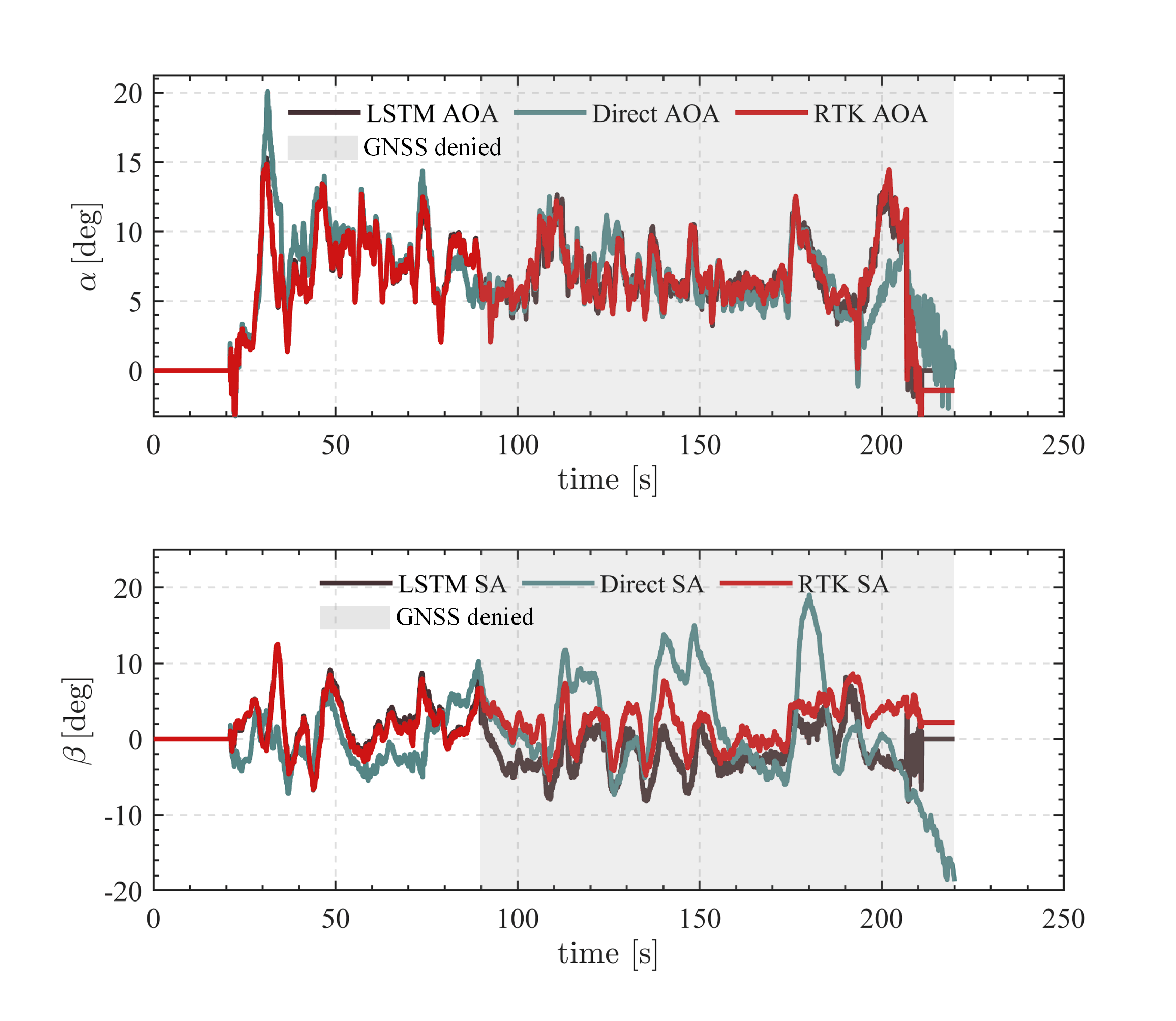}}
\caption{comparison of AOA/SA estimation under GNSS deined.}
\label{AOAerror_GNSSout}
\end{figure}

The Fig. \ref{AOAerror_GNSSout} shows the estimation curves for \( \alpha \) and \( \beta \), with the shaded area representing the GNSS denied. It is evident that at around 190s and 120s, the AOA and the SA calculated by the direct method both exhibit a diverging trend. This is because they rely on velocity measurements, which lack effective constraints when GNSS denied, leading to a significant divergence, especially in the SA.

In contrast, the LSTM prediction network, when integrated with the state estimation system, can effectively maintain estimation accuracy. This shows the strength of using machine learning techniques like LSTM, which can capture the temporal dependencies in the data, to complement traditional state estimation methods, especially when facing sensor outages or failures.

\begin{figure}[tbp]
\centerline{\includegraphics[width=1\linewidth]{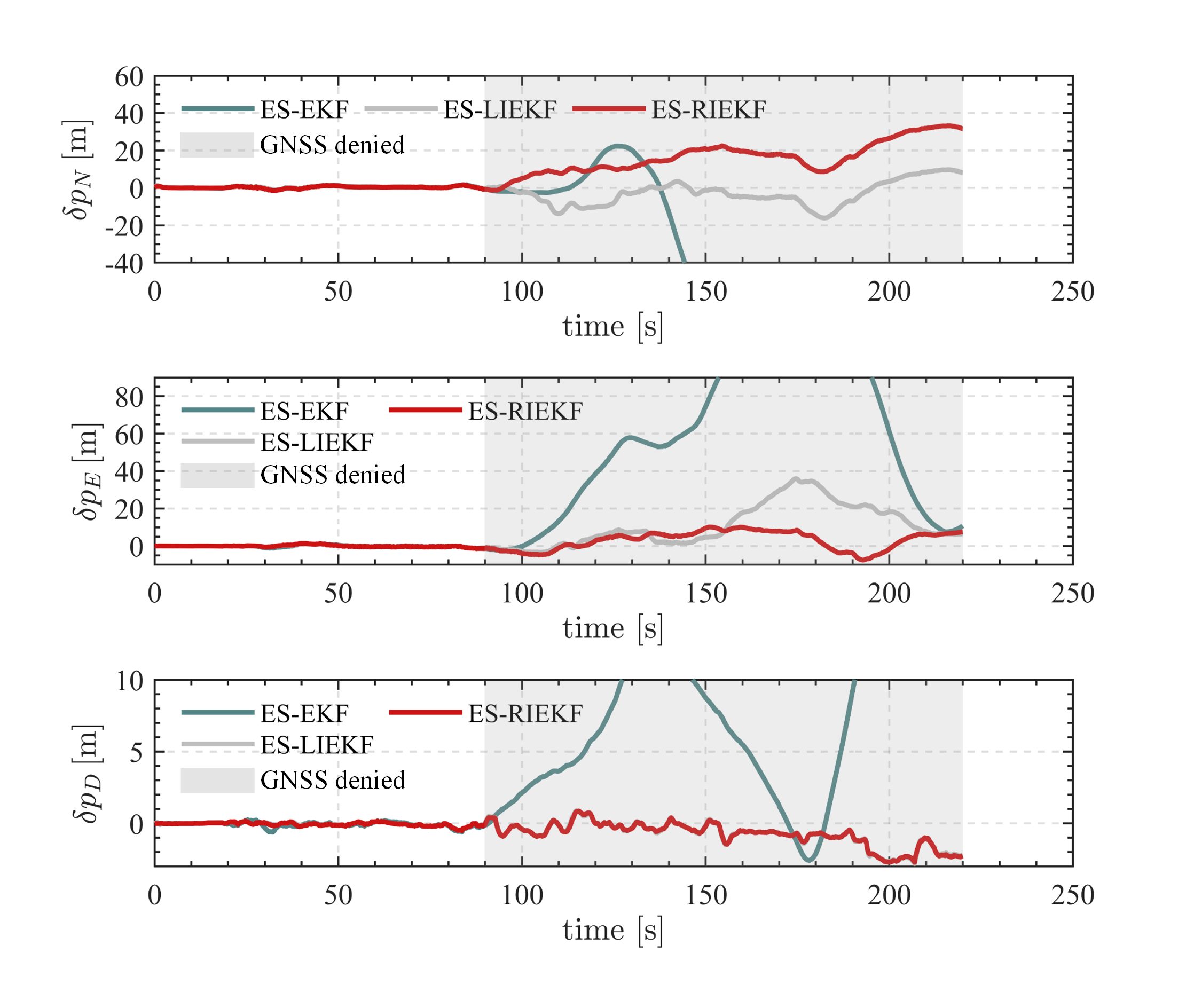}}
\caption{Comparison of position error under GNSS denied.}
\label{poserror_outrange}
\end{figure}

\begin{figure}[tbp]
\centerline{\includegraphics[width=1\linewidth]{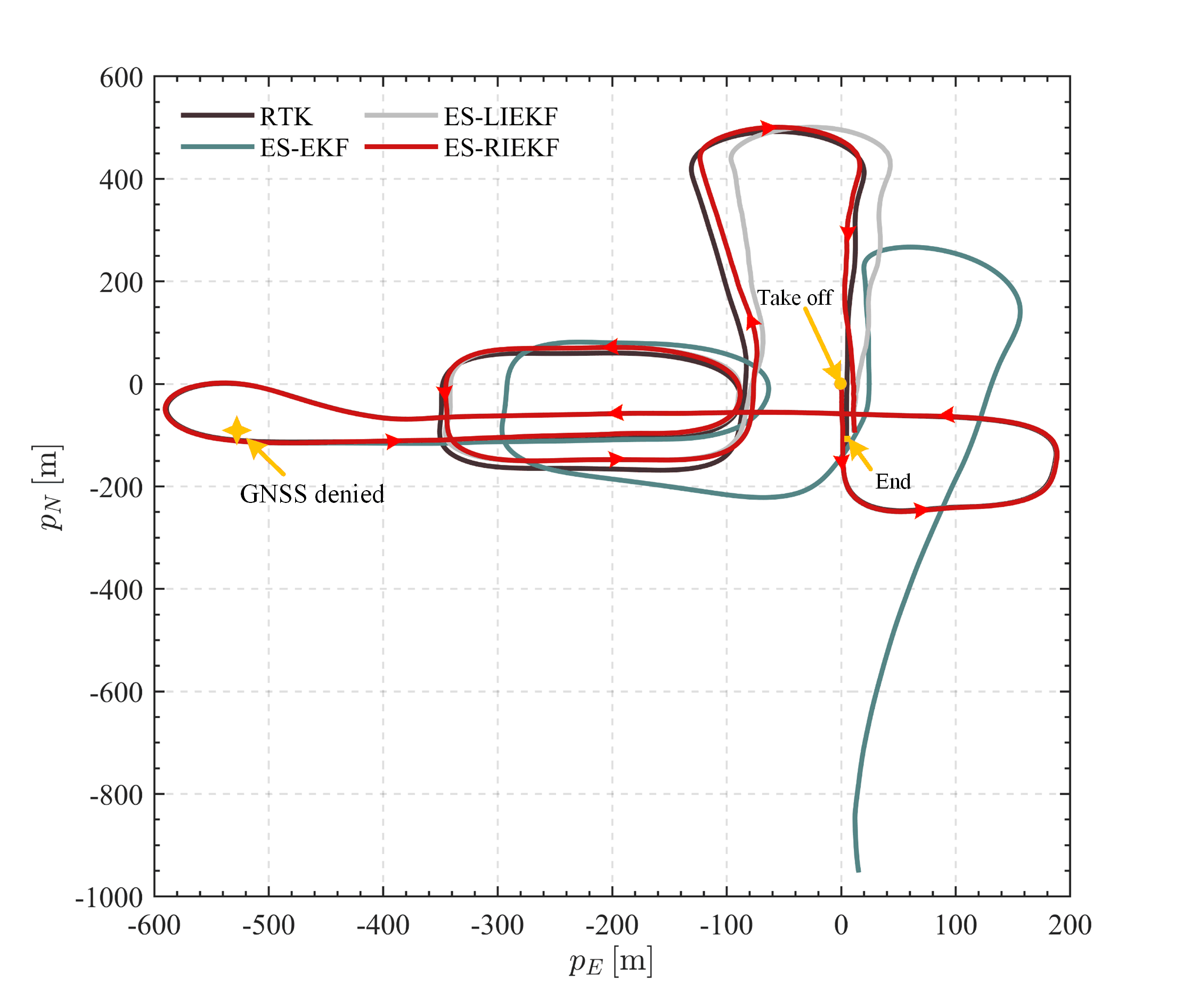}}
\caption{Comparison of 2D-trajectory under GNSS denied.}
\label{2Dtraj_outrange}
\end{figure}

The Fig.\ref{poserror_outrange} shows the position error curves. The north position error is generally slightly larger than the eastward error. Overall, ES-RIEKF has smaller errors, and the vertical error is only about 2m with the constraint of the barometer. Fig. \ref{2Dtraj_outrange} presents the trajectory curves. The yellow stars mark the moment when the denial constraint begins. The green curve represents the pure inertial positioning result after IMU bias correction by the algorithm. The gray and red curves compare the left-invariant and right-invariant algorithms, respectively. It can be seen that under the influence of the 130s denial, the position error of the ES-RIEKF algorithm is within 30m. This significantly improves the accuracy under denial conditions for a fixed-wing UAV with a total journey of approximately 2.5km.
\section{CONCLUSION}\label{section6}
This paper primarily proposes a full-source navigation algorithm for fixed-wing UAVs using ES-RIEKF, demonstrating various advantages of invariant Kalman filtering algorithms under disturbances through real flight data. These advantages include faster convergence and significantly improved accuracy compared to traditional error-state KF. The paper also extends the invariant Kalman filter update equations to constraints like airspeed and control surface deflections, fully utilizing the redundant information from low-cost onboard sensors. It generalizes the invariance framework to AOA and SA, and innovatively establishes an LSTM deep neural network to achieve drift-free prediction of AOA and SA under GNSS denied conditions. This network is further integrated into the ES-RIEKF framework to predict airflow angles when GNSS is unavailable. Additionally, as a resilience advantage inherent to full-source navigation, the paper uses real flight data to demonstrate pose estimation using other sensor combinations when GNSS is denied, thereby effectively suppressing inertial drift through constraints from different sensors. Simulation results show that this approach significantly improves positioning accuracy.

As discussed in the above section, the invariant Kalman is “imperfect” because of the body variables (bias) is not satisfied the state trajectory independent character. In the future work, we need to further focus on new fusion frameworks that can fully integrate the relevant error terms of sensors into the flow-based framework.

\section*{APPENDIX}\label{appendix}
\subsection{Derivation of Invariant Error Compensation}
The exponential matrix map will map an element from $\delta \mathbf{x}$ to $\bchi \in \mathbb{S E}_2(3)$,
\begin{equation}
    \exp{(\delta \mathbf{x}_I)} \triangleq \left[\begin{array}{c|c}
\exp{(\delta \boldsymbol{\theta}_I)} & \mathbf{J}_{l} \left( \delta \boldsymbol{\theta}_I \right) \delta \mathbf{v} \,|\,\mathbf{J}_{l} \left( \delta \boldsymbol{\theta}_I \right) \delta \mathbf{p}  \\
\hline \mathbf{0}_{2\times3} & \mathbf{I}_2
\end{array}\right]
\end{equation}
\begin{equation}
    \exp{(\delta \mathbf{\chi}_{A})} = \exp{(\delta \boldsymbol{\theta}_A)}
\end{equation}
where, $\mathbf{J}_l$ is defined as left Jacobian, writing as \eqref{left Jocobian},
\begin{equation} \label{left Jocobian}
\mathbf{J}_l=\frac{\sin \phi}{\phi} \mathbf{1}+\left(1-\frac{\sin \phi}{\phi}\right) \mathbf{a a}^T+\frac{1-\cos \phi}{\phi} \mathbf{a}^{\wedge}
\end{equation}
where $\phi=|\boldsymbol{\phi}|$ is the angle of rotation and $\mathbf{a}=\boldsymbol{\phi} / \phi$ is the unit-length axis of rotation \cite{barfoot2017state}.

Due to the fact that the bias error and wind error do not satisfy  the $\mathbb{S E}_2(3)$, their compensations are linear.
\begin{equation}
\begin{gathered}
    \mathbf{x}_{b} = \hat{\mathbf{x}}_b+\delta \mathbf{x}_b,\\
    \mathbf{x}_{w} = \hat{\mathbf{v}}^n_{w} + \delta \mathbf{v}^n_{w}
    \end{gathered}
\end{equation}
Above all, the right and left invariant error compensations are expressed as given in \eqref{RInEKF def} and \eqref{left error compensation}, respectively.
\subsection{Derivation of Right-invariant Error Dynamics}\label{appendix B}
Attitude right-invariant error is:
\begin{equation}
    \exp{(\delta \boldsymbol{\theta}^{n \wedge})} = \mathbf{R}^n_b \hat{\mathbf{R}}_b^{n\top}
\end{equation}
Take the derivative of both sides of the equation:
\begin{equation}\label{Rbn dot equ}
    \left[\delta \Dot{\boldsymbol{\theta}}^n \right]_{\times} = \Dot{\mathbf{R}}^n_b \hat{\mathbf{R}}_b^n{^{\top}} +\mathbf{R}^n_b \Dot{\hat{\mathbf{R}}}_b^{n^{\top}}
\end{equation}
according to $ \Dot{\mathbf{R}}^n_b = \mathbf{R}^n_b \left[\omega \right]_{\times}$ and $\exp({\delta \boldsymbol{\theta}^n}) = \mathbf{I}_{3\times3} + \left({\delta \boldsymbol{\theta}^n}\right)_{\times}$ and $\delta \mathbf{b} = \hat{\mathbf{b}} - \mathbf{b}$, (\ref{Rbn dot equ}) can be re-written as \eqref{att err update}:
\begin{table*}
\begin{equation}
\begin{aligned}\label{att err update}
        \left[ \delta \Dot{\boldsymbol{\theta}} \right]^n_{\times} 
    &= \mathbf{R}^n_b \left[\hat{\omega} - \mathbf{b}_g - w_g \right]_{\times} \hat{\mathbf{R}}_b^{n \top}
    + \mathbf{R}^n_b \left( \left[ \hat{\omega} - \hat{\mathbf{b}}_g \right]^{\top}_{\times} \hat{\mathbf{R}}_b^{n^{\top}} \right) = \mathbf{R}^n_b \left(  \left[\hat{\omega}- \mathbf{b}_g - w_g \right]_{\times}-  \left[ \hat{\omega} - \hat{\mathbf{b}}_g \right]_{\times}  \right) \hat{\mathbf{R}}_b^{n^{\top}} \\
    &= \exp(\delta \boldsymbol{\theta}^n) \hat{\mathbf{R}}_b^n\left(  \left[\hat{\omega}- \mathbf{b}_g - w_g \right]_{\times}-  \left[ \hat{\omega} - \hat{\mathbf{b}}_g \right]_{\times}  \right) \hat{\mathbf{R}}_b^{n^{\top}} \approx -\left(\mathbf{I}_{3\times3} + \left({\delta \boldsymbol{\theta}^n}\right)_{\times} \right)\left( \hat{\mathbf{R}}_b^n[\delta \mathbf{b}_g + w_g ]\right)_{\times}
\end{aligned}
\end{equation}
\end{table*}

Ignoring higher-order terms of \(\left[\delta \boldsymbol{\theta}^n \right]_{\times}\) and \(\left[ \delta \mathbf{b}_g\right]_{\times}\), the final expression for the right-invariant attitude error can be derived as follows:
\begin{equation}\label{delta att error update}
     \delta \Dot{\boldsymbol{\theta}}^n = - \hat{\mathbf{R}}_b^n \delta \mathbf{b}_g - \hat{\mathbf{R}}_b^n w_g
\end{equation}

The right-invariant error form for velocity can be represented as follows:
\[
\delta \mathbf{v}^{n} = \mathbf{v}^n - \mathbf{R}_b^n \hat{\mathbf{R}_b^n}^{\top} \mathbf{\hat{v}}^n
\]
Differentiating both sides of  \eqref{state update dynamic}:
\begin{equation}\label{dot v err update}
    \begin{aligned}
        \delta \Dot{ \mathbf{v}}^{n} = \Dot{\mathbf{v}}^n 
        - \Dot{\mathbf{R}}^n_b  \hat{\mathbf{R}}_b^{n \top} \hat{\mathbf{v}}^n 
        - \mathbf{R}^n_b \Dot{\hat{\mathbf{R}}}^n_b\hat{\mathbf{v}}^n -  \mathbf{R}^n_b\hat{\mathbf{R}_b^n}^{\top} \Dot{ \hat{\mathbf{v}}}^{n}
    \end{aligned}
\end{equation}
According to \eqref{state update dynamic}, simplifying \eqref{dot v err update}, we can get  right invariant velocity error eqution \eqref{delta v error update}. 

\begin{table*}[htbp]
\begin{equation}\label{delta v error update}
    \begin{aligned}
        \delta \Dot{ \mathbf{v}}^{n} &= 
        \underbrace{{\exp{(\delta \boldsymbol{\theta}^n)}}\hat{\mathbf{R}_b^n} \left( \hat{\mathbf{a}}^b - \mathbf{b}_a - w_a \right) + \mathbf{g}^n}_{\Dot{\mathbf{v}^n}} 
        - \underbrace{\exp{(\delta \boldsymbol{\theta}^n})\hat{\mathbf{R}_b^n} \left[ \hat{\mathbf{\omega}^b} - \mathbf{b}_g - w_g \right]_{\times}\hat{\mathbf{R}_b^n}^{\top} \hat{\mathbf{v}}^n}_{\Dot{\mathbf{R}}^n_b  \hat{\mathbf{R}_b^n}^{\top} \hat{\mathbf{v}}^n} \\
        &- \underbrace{\exp{(\delta \boldsymbol{\theta}^n)}\mathbf{R}^n_b \left[\hat{\mathbf{\omega}^b} - \hat{\mathbf{b}_g} \right]_{\times} \hat{\mathbf{R}_b^n}^{\top} \hat{\mathbf{v}}^n
        }_{\mathbf{R}^n_b \Dot{\hat{\mathbf{R}}}^n_b\hat{\mathbf{v}}^n}
        - \underbrace{\exp{(\delta \boldsymbol{\theta}^n)}\left[ \hat{\mathbf{R}}^n_b \left(\hat{\mathbf{a}^b} - \hat{\mathbf{b}}_a  \right) + \mathbf{g}^n\right]
        }_{\mathbf{R}^n_b\hat{\mathbf{R}_b^n}^{\top} \Dot{ \hat{\mathbf{v}}}^{n}}\\
        &= \exp{(\delta \boldsymbol{\theta}^n)}\left[ \hat{\mathbf{R}}^n_b\left( \hat{\mathbf{b}_a} - \mathbf{b}_a - w_a \right)\right] + \mathbf{g}^n \left( \boldsymbol{I}- \exp{(\delta \boldsymbol{\theta}^n)} \right) - \exp{(\delta \boldsymbol{\theta}^n)} \left( \hat{\mathbf{R}}_b^n[ \hat{\mathbf{b}}_g - \mathbf{b}_g + w_g ]\right)_{\times}\mathbf{v}^n\\
        &\approx -\hat{\mathbf{R}}^n_b\left( \delta\mathbf{b}_a - w_a \right)  + [\mathbf{g}^n]_{\times} \delta \boldsymbol{\theta}^n - \mathbf{v}^n \hat{\mathbf{R}}_b^n\left( \delta{\mathbf{b}}_g + w_g \right)\\
        &= [\mathbf{g}^n]_{\times}\delta \boldsymbol{\theta}^n - \hat{\mathbf{R}}_b^n \delta \mathbf{b}_a - [\mathbf{v}^n]_{\times} \hat{\mathbf{R}}_b^n\delta{\mathbf{b}}_g  - \hat{\mathbf{R}}_b^n w_a - [\mathbf{v}^n]_{\times} \hat{\mathbf{R}}_b^n w_g 
    \end{aligned}
\end{equation}
\end{table*}

The right-invariant position error \( \delta \mathbf{p}^n \) can be represented as follows:
\begin{equation}\label{delta p err equ}
    \delta \mathbf{p}^n = \mathbf{p}^n - \mathbf{R}_b^n {\hat{\mathbf{R}}}_b^{n\top} \mathbf{\hat{p}}^n
\end{equation}

According to \( \Dot{\mathbf{p}}^n = \mathbf{v}^n+\boldsymbol{\mathbf{R}}^n_b \mathbf{a}^b \delta t + \mathbf{g}^n \delta t/2 \), (\ref{delta p err equ}) can be rewritten as \eqref{delta p error equ}.
\begin{table*}
\begin{equation}\label{delta p error equ}
    \begin{aligned}
        \delta \Dot{\mathbf{p}}^n &= \Dot{\mathbf{p}}^n 
        - \Dot{\mathbf{R}}^n_b  \hat{\mathbf{R}_b^n}^{\top} \hat{\mathbf{p}}^n 
        - \mathbf{R}^n_b \Dot{\hat{\mathbf{R}}}^n_b\hat{\mathbf{p}}^n -  \mathbf{R}^n_b\hat{\mathbf{R}_b^n}^{\top} \Dot{ \hat{\mathbf{p}}}^{n}\\
        &= \underbrace{\exp{(\delta \boldsymbol{\theta}^n)} \hat{\mathbf{v}}^n + \delta \mathbf{v}^n+\exp{(\delta \boldsymbol{\theta})^n} \hat{\boldsymbol{\mathbf{R}}}^n_b \left(\hat{\mathbf{a}}^b - \mathbf{b}_g - w_a \right) \delta t + \mathbf{g}^n \delta t/2}_{\Dot{\mathbf{p}}^n} 
        - \underbrace{\exp{(\delta \boldsymbol{\theta}^n})\hat{\mathbf{R}_b^n} \left[ \hat{\mathbf{\omega}^b} - \mathbf{b}_g - w_g \right]_{\times}\hat{\mathbf{R}_b^n}^{\top} \hat{\mathbf{p}}^n}_{\Dot{\mathbf{R}}^n_b  \hat{\mathbf{R}_b^n}^{\top} \hat{\mathbf{p}}^n} \\
        &- \underbrace{\exp{(\delta \boldsymbol{\theta}^n)}\mathbf{R}^n_b \left[\hat{\mathbf{\omega}^b} - \hat{\mathbf{b}_g} \right]_{\times} \hat{\mathbf{R}_b^n}^{\top} \hat{\mathbf{p}}^n
        }_{\mathbf{R}^n_b \Dot{\hat{\mathbf{R}}}^n_b\hat{\mathbf{p}}^n}
        - \underbrace{\exp{(\delta \boldsymbol{\theta}^n)} \left[ \hat{\mathbf{v}}^n+\hat{\boldsymbol{\mathbf{R}}}^n_b \left( \hat{\mathbf{a}}^b - \hat{\mathbf{b}}_a \right)\delta t + \mathbf{g}^n \delta t/2\right]
        }_{\mathbf{R}^n_b\hat{\mathbf{R}_b^n}^{\top} \Dot{ \hat{\mathbf{p}}}^{n}}\\
        &= \delta \mathbf{v}^n + \exp{(\delta \boldsymbol{\theta}^n)} \hat{\boldsymbol{\mathbf{R}}}^n_b \left( \hat{\mathbf{b}}_a  - \mathbf{b}_a - w_a\right) \delta t + \mathbf{g}^n \delta t/2 \left( \boldsymbol{I}- \exp{(\delta \boldsymbol{\theta}^n)} \right) \\
        &- \exp{(\delta \boldsymbol{\theta}^n)} \left( \hat{\mathbf{R}}_b^n[ \hat{\mathbf{b}}_g - \mathbf{b}_g + w_g ]\right)_{\times}\mathbf{p}^n\\
        &\approx \delta \mathbf{v}^n + [ \mathbf{g}^n \delta t/2 ]_{\times} \delta \boldsymbol{\theta}^n -  \hat{\mathbf{R}}_b^n \delta \mathbf{b}_a - [\mathbf{p}^n]_{\times} \hat{\mathbf{R}}_b^n\delta{\mathbf{b}}_g  - \hat{\mathbf{R}}_b^n w_a - [\mathbf{p}^n]_{\times} \hat{\mathbf{R}}_b^n w_g 
    \end{aligned}
\end{equation}
\end{table*}

\subsection{Derivation of Left-invariant Error Dynamics}\label{appendix C}
The definition of the left-invariant error state variables is the same as in RI-EKF.
\begin{equation}
\delta \mathbf{x}=\left[\begin{array}{lllllll}
\delta \boldsymbol{\theta}_I^{n \top} & 
\delta \mathbf{p}^{n \top} & 
\delta \mathbf{v}^{n \top} & 
\delta \boldsymbol{\theta}_w^{n \top} & 
\delta \bm{v}^{w \top}_{wind}
\end{array}\right]^{\top}
\end{equation}
The relationship between the state left error and the estimated value can be described as follows:
\begin{equation}\label{left error compensation}
\begin{aligned}
\mathbf{\chi} &= \hat{\mathbf{\chi}} \boxplus \delta \mathbf{x}\\
&=
\left( \hat{\mathbf{x}}_{I} \exp \left(\delta \mathbf{\chi}_{I}\right) ,
\hat{\mathbf{x}}_b+\delta \mathbf{x}_b,
 \hat{\mathbf{x}}_{A} \exp \left(\delta \mathbf{\chi}_{A}\right),
\hat{\mathbf{x}}_{vw}+\delta \mathbf{x}_{vw}
\right) \\
 &= \left[\begin{array}{c}
    \mathbf{R}_b^n \exp(\delta \boldsymbol{\theta}_I)\\
   \mathbf{\hat{p}}^n +\mathbf{R}_b^n \mathbf{J}_l(\delta \boldsymbol{\theta}_I))\delta \mathbf{p} \\
   \mathbf{\hat{v}}^n  + \mathbf{R}_b^n \mathbf{J}_l(\delta \boldsymbol{\theta}_I)) \delta \mathbf{v} \\
   \mathbf{\Hat{b}}_g + \delta \mathbf{b}_g \\
   \mathbf{\Hat{b}}_a + \delta \mathbf{b}_a \\
    \mathbf{R}_a^b \exp(\delta \boldsymbol{\theta}_a)\\
   \hat{\mathbf{v}}^n_{w} + \delta \mathbf{v}^n_{w}\\
\end{array}\right]
\end{aligned}
\end{equation}

From the above equation, we know:
\[
\exp (\delta \boldsymbol{\theta}^n_L) = \hat{\mathbf{R}}^b_n \mathbf{R}^n_b = \hat{\mathbf{R}}^b_n \exp (\delta \boldsymbol{\theta}^n_R) \hat{\mathbf{R}}^n_b
\]
And the following approximation holds:
\[
\exp (\delta \boldsymbol{\theta}^n) \approx \boldsymbol{I}_3 + [\boldsymbol{\theta}^n]_{\times}
\]
Then, the left-invariant angular error rate equation can be expressed as:
\begin{equation}\label{LI_att}
 \delta \Dot{\boldsymbol{\theta}}^n_L = -[\omega]_{\times} \delta \boldsymbol{\theta}^n_L - \delta \boldsymbol{b}_g - w_g   
\end{equation}
represents the dynamics equation for right-invariant attitude error. 

As for the dynamics equation for left-invariant velocity error, it is given by:
\begin{equation}
    \delta \mathbf{v}^n_L = \hat{\mathbf{R}}^{n \top}_b \left( \mathbf{v}^n - \hat{\mathbf{v}}^n \right)
\end{equation}
Taking the derivative of both sides of the equation, we obtain:
\begin{equation} \label{vn LI 1}
    \begin{aligned}
        \delta \Dot{\mathbf{v}}^n_L &= \Dot{\hat{\mathbf{R}}}^{n \top}_b \left( \mathbf{v}^n - \hat{\mathbf{v}}^n \right)
        + \hat{\mathbf{R}}^{n \top}_b \left( \Dot{\mathbf{v}}^n - \Dot{\hat{\mathbf{v}}}^n \right)
    \end{aligned}
\end{equation}
According to $\Dot{\hat{\mathbf{R}}}^{n \top}_b = -[\omega]_{\times}{\hat{\mathbf{R}}}^{n \top}_b$,$\Dot{\mathbf{v}} = \mathbf{R}^n_b \mathbf{a}^b + \mathbf{g}^n$,$\mathbf{R}^b_n = \hat{\mathbf{R}}^b_n \exp (\delta \boldsymbol{\theta}^n_L)$, (\ref{vn LI 1}) can be simplified as \eqref{left v eq}.
\begin{table*}
\begin{equation}\label{left v eq}
    \begin{aligned}
        \delta \Dot{\mathbf{v}}^n_L &= -[\boldsymbol{\omega}]_{\times}  \delta \mathbf{v}^n_L + 
        {\hat{\mathbf{R}}}^{n \top}_b \left( \underbrace{{\exp{(\delta \boldsymbol{\theta}^n)}}\hat{\mathbf{R}_b^n} \left( \hat{\mathbf{a}}^b - \mathbf{b}_a - w_a \right) + \mathbf{g}^n}_{\Dot{\mathbf{v}^n}} -
         \underbrace{\hat{\mathbf{R}_b^n} \left( \hat{\mathbf{a}}^b - \hat{\mathbf{b}}_a \right) - \mathbf{g}^n}_{\Dot{\hat{\mathbf{v}}}^n} 
        \right) \\
        &= -[\hat{\mathbf{a}}^b]_{\times} \delta \boldsymbol{\theta}^n_L - [\boldsymbol{\omega}^b]_{\times} \delta \mathbf{v}^n_L - \delta \mathbf{b}_a - w_a
    \end{aligned}
\end{equation}
\end{table*}
Similarly, the left-invariant position error equation is given by:
\begin{equation}
    \delta \mathbf{p}^n_L = \hat{\mathbf{R}}^{n \top}_b \left( \mathbf{p}^n - \hat{\mathbf{p}}^n \right)
\end{equation}
With $\Dot{\mathbf{p}}^n = \mathbf{v}^n$, the equation simplifies to:
\begin{equation}
\begin{aligned}
    \delta \Dot{\mathbf{p}}^n_L &= -[\boldsymbol{\omega}]_{\times}\delta \mathbf{p}^n_L + \hat{\mathbf{R}}^{n \top}_b \left(\Dot{\mathbf{p}}^n - \Dot{\hat{\mathbf{p}}}^n \right)\\
    &= -[\boldsymbol{\omega}]_{\times}\delta \mathbf{p}^n_L + \delta \mathbf{v}^n_L
\end{aligned}
\end{equation}

Above all, the $\mathbf{F}^L_{davp}$ matrix can be represented as follows:
\begin{equation}\label{F_davp}
    \mathbf{F}^L_{davp} = \left[ \begin{array}{ccc}
         -[\boldsymbol{\omega}]_{\times} & \boldsymbol{0}_{3\times3} & \boldsymbol{0}_{3\times3}\\
        \boldsymbol{0}_{3\times3} & -[\boldsymbol{\hat{a}^b}]_{\times} & -[\boldsymbol{\omega}]_{\times} \\
        \boldsymbol{0}_{3\times3} & \boldsymbol{I}_{3\times3}  &-[\boldsymbol{\omega}]_{\times} \\
    \end{array} \right]_{9\times9}
\end{equation}
The matrix $\mathbf{F}^L_{davp2b}$ relating the AVP error states to the bias error states $\delta \mathbf{b}_g$ and $\delta \mathbf{b}_a$ can be expressed as follows:
\begin{equation}
\mathbf{F}^L_{davp2b} = \left[
    \begin{array}{cc}
       - \boldsymbol{I}_{3\times3} &  \boldsymbol{0}_{3\times3}\\
        \boldsymbol{0}_{3\times3}  & - \boldsymbol{I}_{3\times3} \\
        \boldsymbol{0}_{3\times3} & \boldsymbol{0}_{3\times3} \\
    \end{array} \right]_{9\times6}
\end{equation}
\section*{ACKNOWLEDGMENT}

Thanks to Yan Hongtao, Xu Yong and others for providing a lot of help in the flight experiment.

\bibsection*{REFERENCES}
\def\refname{\vskip -0.5cm}
\bibliographystyle{IEEEtran}
\bibliography{reference}



\end{document}